%% file: main.tex
\documentclass[11pt]{article}

\usepackage{fullpage}
\usepackage[ruled, linesnumbered, vlined, commentsnumbered]{algorithm2e}
\usepackage{graphicx,subfig} 
\usepackage{amsfonts,amssymb,amsmath,amsthm,amsopn}	
\usepackage{hhline,booktabs,colortbl,multirow,tabularx,threeparttable} 

\usepackage{enumerate}
\usepackage{authblk}
\usepackage{footnote}
\usepackage{hyperref}
\usepackage{prettyref}
\usepackage{cite}
\usepackage{setspace}
\usepackage{color}
\usepackage{xcolor}  
\usepackage{scrpage2}

\usepackage{tikz}
\usepackage{pgfplots}
\usetikzlibrary{positioning,shapes,shadows,arrows,calc}
\tikzstyle{component}=[rectangle, draw=black, rounded corners, fill=blue!40, drop shadow, text centered, anchor=north, text=white, minimum height=1cm]
\tikzstyle{arrow}=[->, thick]

\pgfplotsset{compat=1.12}
\usetikzlibrary{intersections}
\usetikzlibrary{pgfplots.statistics}
\usepgfplotslibrary{fillbetween}

\definecolor{myblue}{RGB}{34,31,217}
\definecolor{mycyan}{gray}{.7}
\definecolor{Gray}{gray}{0.9}

\newcommand{\pref}{\prettyref}

\newrefformat{fig}{Fig.~\ref{#1}}
\newrefformat{tab}{Table~\ref{#1}}
\newrefformat{sec}{Section~\ref{#1}}
\newrefformat{app}{Appendix~\ref{#1}}
\newrefformat{alg}{Algorithm~\ref{#1}}
\newrefformat{property}{Property~\ref{#1}}
\newrefformat{theorem}{Theorem~\ref{#1}}
\newrefformat{corollary}{Corollary~\ref{#1}}
\newrefformat{proposition}{Proposition~\ref{#1}}
\newrefformat{def}{Definition~\ref{#1}}
\newrefformat{eq}{equation~(\ref{#1})}

\renewcommand\Authands{ and }

\usepackage{lscape}

\begin{document}

\title{\vspace{-3em}\LARGE\textbf{Interactive Decomposition Multi-Objective Optimization via Progressively Learned Value Functions}\footnote{This paper is submitted for possible publication. Reviewers can use this manuscript as an alternative in peer review.}}

\author[1]{\normalsize Ke Li$^{\#}$}
\author[2]{\normalsize Renzhi Chen$^{\#}$}
\author[3]{\normalsize Dragan Savi\'c}
\author[2]{\normalsize Xin Yao}
\affil[1]{\normalsize Department of Computer Science, University of Exeter}
\affil[2]{\normalsize CERCIA, School of Computer Science, University of Birmingham}
\affil[3]{\normalsize Department of Engineering, University of Exeter}
\affil[$\ast$]{\normalsize Email: \texttt{\{k.li, d.savic\}@exeter.ac.uk, \{rxc332, x.yao\}@cs.bham.ac.uk}}
\affil[$\#$]{\normalsize The first two authors make equal contributions to this paper.}

\renewcommand\Authands{ and }

\date{}
\maketitle

{\normalsize\textbf{Abstract: } } Decomposition has become an increasingly popular technique for evolutionary multi-objective optimization (EMO). A decomposition-based EMO algorithm is usually designed to approximate a whole Pareto-optimal front (PF). However, in practice, the decision maker (DM) might only be interested in her/his region of interest (ROI), i.e., a part of the PF. Solutions outside that might be useless or even noisy to the decision-making procedure. Furthermore, there is no guarantee to find the preferred solutions when tackling many-objective problems. This paper develops an interactive framework for the decomposition-based EMO algorithm to lead a DM to the preferred solutions of her/his choice. It consists of three modules, i.e., consultation, preference elicitation and optimization. Specifically, after every several generations, the DM is asked to score a few candidate solutions in a consultation session. Thereafter, an approximated value function, which models the DM's preference information, is progressively learned from the DM's behavior. In the preference elicitation session, the preference information learned in the consultation module is translated into the form that can be used in a decomposition-based EMO algorithm, i.e., a set of reference points that are biased toward to the ROI. The optimization module, which can be any decomposition-based EMO algorithm in principle, utilizes the biased reference points to direct its search process. Extensive experiments on benchmark problems with three to ten objectives fully demonstrate the effectiveness of our proposed method for finding the DM's preferred solutions.

{\normalsize\textbf{Keywords: } }Multi-criterion decision making, interactive multi-objective optimization, decomposition-based technique, evolutionary computation.

\input{introduction}

\input{preliminary}

\input{proposal}

\input{settings}

\input{experiments}

\input{conclusion}

\section*{Acknowledgment}
This work was supported by the Ministry of Science and Technology of China (Grant No. 2017YFC0804002), the Science and Technology Innovation Committee Foundation of Shenzhen (Grant No. ZDSYS201703031748284) and EPSRC (Grant No. EP/J017515/1).

\bibliographystyle{IEEEtran}
\bibliography{IEEEabrv,preference}

\end{document}

%% file: introduction.tex

\section{Introduction}
\label{sec:introduction}

Multi-objective optimization problems (MOPs) involve optimizing more than one objective function simultaneously. They typically arise in various fields of science (e.g.,~\cite{PonsichJC13,AntonelliDM14,QasemSHDA13}) and engineering (e.g.~\cite{RostamiOSB14,MontanoCM12,LogisticApp}) where optimal decisions need to be taken in the presence of trade-offs between two or more conflicting objectives. For example, in portfolio management, maximizing the expected value of portfolio returns and minimizing the potential risk are two typical conflicting objectives.

Due to the population-based property, evolutionary algorithms (EAs) have been widely recognized as a major approach for multi-objective optimization. Over the last three decades and beyond, much effort has been dedicated to developing evolutionary multi-objective optimization (EMO) algorithms~\cite{ZhouZLL09,LiZLZL09,LiKWCR12,LiKCLZS12,LiKWTM13,IJUFKS13,LiZKLW14,LiFKZ14,LiK14,WuKZLWL15,LiDZ15,LiDZK15,LiKD15,LiODY16,LiD16,WuKJLZ17,LiDAY17,WuLKZZ17,LiDZZ17,ChenLY18,LiDY18,LiCFY18,WuLKZZ18,LiCMY18}, such as non-dominated sorting genetic algorithm II (NSGA-II)~\cite{DebAPM02}, improved strength Pareto EA (SPEA2)~\cite{ZitzlerLT01} and multi-objective EA based on decomposition (MOEA/D)~\cite{ZhangL07}, to find a set of well-converged and well-diversified efficient solutions that approximate the whole Pareto-optimal front (PF). Nevertheless, the ultimate goal of MO is to help the decision maker (DM) find solutions that meet at most her/his preferences. Supplying a DM with a large amount of widely spread trade-off alternatives not only increases her/his workload, but also provides many irrelevant or even noisy information to the decision-making procedure. Moreover, due to the curse of dimensionality, the performance of EMO algorithms degenerate with the number of objectives. In addition, the number of points to represent a PF grows exponentially with the number of objectives, thereby requiring a large population size to run an EMO algorithm. Besides, there is a severe cognitive obstacle for the DM to comprehend a high-dimensional PF.

To alleviate the above problems associated with the \textit{a posteriori} decision-making procedure in the traditional EMO, it is more practical to incorporate the DM's preference information into the search process. This allows the computational efforts to concentrate on the region of interest (ROI) and thus has a better approximation therein. In general, the preference information can be incorporated \textit{a priori} or \textit{interactively}. If the preference information (in the form of one or more reference points, reference directions or light beams) is elicited \textit{a priori}, it is used to guide the population toward the ROI. For example, the cone-domination based EMO~\cite{BrankeKS01}, biased niching based EMO~\cite{Deb03,BrankeD05}, reference point based EMO~\cite{DebSBC06}, the reference direction based EMO~\cite{DebK07GECCO} and the light beam based EMO~\cite{DebK07CEC} are attempts along this direction. Moreover, in~\cite{GreenwoodHD96}, Greenwood~\textit{et al}. derived a linear value function from a given ranking of a few alternatives to model the DM's preference information. Thereafter, this linear value function is used as the fitness function in an EMO algorithm to guide the population toward the ROI. Note that, in the \textit{a priori} approach, the DM only interacts at the beginning of an EMO run. However, it is non-trivial to faithfully represent the preference information before solving the MOP at hand.

In practice, eliciting the preference information in an interactive manner, which has been studied in the multi-criterion decision-making (MCDM) field for over half a century, seems to be more interesting. This enables the DM to progressively learn and understand the characteristics of the MOP at hand and adjust their elicited preference information. Consequently, solutions are effectively driven toward the ROI. In principle, the above mentioned \textit{a priori} EMO approaches can also be used in an interactive EMO approach in an iterative manner (e.g.,~\cite{DebK07GECCO} and~\cite{DebK07CEC}). Specifically, in the first round, the DM can elicit certain preference information (in the form of reference points, reference directions or other means), and it is used in an EMO algorithm to find a set of preferred Pareto-optimal solutions. Thereafter, a few representative solutions can be shown to the DM. If these solutions are satisfied by the DM, they will be used as the outputs and the iterative procedure terminates. Otherwise, the DM will adjust her/his preference information accordingly and it will be used in another EMO run. Alternatively, the DM can be involved to periodically provide preference information as the EMO iterations are underway~\cite{PhelpsK03}. In particular, the preference information is progressively learned as value functions with the evolution of solutions. Since the DM gets more frequent chance to provide new information, the overall process is more DM-oriented. Moreover, the DM may feel more in charge and more involved in the overall \textit{optimization-cum-decision-making} process. 

During recent years, especially after the developments of MOEA/D and NSGA-III~\cite{DebJ14}, the decomposition-based EMO methods have become increasingly popular for the \textit{a posteriori} MO. Generally speaking, by specifying a set of reference points\footnote{In this paper, we use the term reference point without loss of generality, although some other papers, e.g, the original MOEA/D~\cite{ZhangL07}, also use the term weight vector interchangeably.}, the decomposition-based EMO methods at first decompose the MOP at hand into multiple subproblems, either with scalar objective or simplified multi-objective. Then, a population-based technique is applied to solve these subproblems in a collaborative manner. Under some mild conditions, the optimal solutions of all subproblems constitute a good approximation to the PF. It is not difficult to understand that the distribution of the reference points is essential for a decomposition-based EMO method. It not only implies \textit{a priori} assumption of the PF's geometrical characteristics, but also determines the distribution of Pareto-optimal solutions. There have been some studies on how to generate desired reference points. For example, \cite{NBI} suggested a structured method to generate evenly distributed reference points on a canonical simplex. To adapt to the irregular PFs, such as disconnected or mixed shapes and disparately scaled objectives, some adaptive reference point adjustment methods (e.g., \cite{JiangCZO11} and \cite{QiMLJSW14}) have been developed to progressively adjust the distribution of reference points on the fly. To integrate the DM's preference information into the decomposition-based EMO methods, a natural idea is to make the distribution of the reference points be biased toward the ROI. Although it sounds quite intuitive, in practice, how to obtain the appropriate reference points that accommodate to the DM's preference information is far from trivial. Most recently, there have been some limited initiatives on adjusting the distribution of the reference points according to the DM's preference information (e.g.,~\cite{MohammadiOLD14} and~\cite{MaLQLJDWDHZW16}). However, most, if not all, of them specify the DM's preference information in \textit{a priori} manner.

This paper develops a simple interactive framework for the decomposition-based EMO algorithms that can progressively learn an approximated value function (AVF) from the DM's behavior and thus guide the population toward the ROI. This framework consists of three interdependent modules: optimization, consultation and preference elicitation. Specifically, 
\begin{itemize}
    \item The optimization module can be any decomposition-based EMO algorithm in principle. It uses the preference information elicited from the preference elicitation module to find the preferred solutions. Periodically, it supplies the consultation module with a few candidates for learning an AVF.
    \item The consultation module is the interface by which the DM interacts with the optimization module. It simulates the DM that assigns a score to each candidate supplied by the optimization module. Then, by using the scored candidates found so far as the training data, a machine learning algorithm is applied to find an AVF that models the DM's preference information.
    \item The preference elicitation module aims at translating the preference information learned from the consultation module in the form that can be used in a decomposition-based EMO algorithm. In particular, it changes the distribution of reference points to be focused in the ROI.
\end{itemize}
In empirical studies, our proposed interactive framework is embedded in two widely used decomposition-based EMO algorithms, i.e., MOEA/D and NSGA-III. Their effectiveness for finding preferred Pareto-optimal solutions are validated on several benchmark problems with three to ten objectives.

The rest of this paper is organized as follows. \pref{sec:preliminaries} provides some preliminaries of this paper. \pref{sec:proposal} describes the technical details of the interactive framework step by step. Afterwards, in~\pref{sec:settings} and~\ref{sec:results}, the effectiveness of the proposed method is empirically investigated on various benchmark problems with three to ten objectives. \pref{sec:conclusions} concludes this paper and provides some future directions.

%% file: preliminary.tex

\section{Preliminaries}
\label{sec:preliminaries}

In this section, we first provide some basic definitions of MO. Then, to facilitate the descriptions of our proposed interactive framework for the decomposition-based EMO algorithms, we start from describing the working mechanisms of two widely used decomposition-based EMO algorithms, i.e., MOEA/D and NSGA-III. At the end, we briefly overview the past studies of interactive MO.

\subsection{Basic Definitions}
\label{sec:definitions}

The MOP considered in this paper is formulated as:
\begin{equation}
\begin{array}{l}
\mathrm{minimize} \quad \mathbf{F}(\mathbf{x})=(f_1(\mathbf{x}),\cdots,f_m(\mathbf{x}))^{T}\\
\mathrm{subject\ to} \quad \mathbf{x}\in\Omega
\end{array},
\label{eq:MOP}
\end{equation}
where $\mathbf{x}=(x_1,\cdots,x_n)^T$ is a $n$-dimensional decision vector and $\mathbf{F}(\mathbf{x})$ is an $m$-dimensional objective vector. $\Omega$ is the feasible set in the decision space $\mathbb{R}^n$ and $\mathbf{F}:\Omega\rightarrow\mathbb{R}^m$ is the corresponding attainable set in the objective space $\mathbb{R}^m$. Without considering the DM's preference information, given two solutions $\mathbf{x}^1,\mathbf{x}^2\in\Omega$, $\mathbf{x}^1$ is said to dominate $\mathbf{x}^2$ if and only if $f_i(\mathbf{x}^1)\leq f_i(\mathbf{x}^2)$ for all $i\in\{1,\cdots,m\}$ and $\mathbf{F}(\mathbf{x}^1)\neq\mathbf{F}(\mathbf{x}^2)$. A solution $\mathbf{x}\in\Omega$ is said to be Pareto-optimal if and only if there is no solution $\mathbf{x}^\prime\in\Omega$ that dominates it. The set of all Pareto-optimal solutions is called the Pareto-optimal set (PS) and their corresponding objective vectors form the PF. Accordingly, the ideal point is defined as $\mathbf{z}^{\ast}=(z^{\ast}_1,\cdots,z^{\ast}_m)^T$, where $z^{\ast}_i=\min\limits_{\mathbf{x}\in PS}f_i(\mathbf{x})$, and the nadir point is defined as $\mathbf{z}^{nd}=(z^{nd}_1,\cdots,z^{nd}_m)^T$, where $z^{nd}_i=\max\limits_{\mathbf{x}\in PS}f_i(\mathbf{x})$, $\forall i\in\{1,\cdots,m\}$.

\subsection{Decomposition-based EMO algorithms}
\label{sec:DEMO}

\subsubsection{MOEA/D}
The basic idea of MOEA/D is to decompose the original MOP into several subproblems and it uses a population-based technique to solve these subproblems in a collaborative manner. In particular, with respect to a reference point $\mathbf{w}$, this paper uses the Tchebycheff function~\cite{LiZKLW14,LiKZD15} as a subproblem which is defined as:
\begin{equation}
    \begin{aligned}
        \mathrm{minimize~} g(\mathbf{x}|\mathbf{w},\mathbf{z}^\ast)&=\max_{1\leq i\leq m}{|f_i(\mathbf{x})-z_i^\ast|/w_i}\\
        \mathrm{subject\ to}\quad \mathbf{x}\in\Omega
    \end{aligned}
    \label{eq:TCH}
\end{equation} 
where $\mathbf{z}^\ast$ is the ideal point. The general working mechanisms of MOEA/D is given as the following three-step process.
\begin{enumerate}[Step 1:]
    \item Initialize a population of solutions $P:=\{\mathbf{x}^i\}_{i=1}^N$, a set of reference points $W:=\{\mathbf{w}^i\}_{i=1}^N$ and their neighborhood structure. Randomly assign each solution to a reference point.
    \item For $i=1,\cdots,N$, do
        \begin{enumerate}[Step 2.1:]
            \item Randomly selects a required number of mating parents from $\mathbf{w}^i$'s neighborhood.
            \item Use crossover and mutation to reproduce offspring $\mathbf{x}^c$.
            \item Update the subproblems within the neighborhood of $\mathbf{w}^i$ by $\mathbf{x}^c$.
        \end{enumerate}
    \item If the stopping criteria is met, then stop and output the population. Otherwise, go to Step 2.
\end{enumerate}

We would like to make some remarks on some important ingredients of the above MOEA/D procedure.
\begin{itemize}
    \item In Step 1, we use the classic method developed by Das and Dennis~\cite{NBI} to initialize a set of evenly distributed reference points from a canonical simplex. Furthermore, the neighborhood structure $B(i)$ of each reference point $\mathbf{w}^i$, $i\in\{1,\cdots,N\}$, contains its $T$ closest reference points, where $T=20$ as suggested in~\cite{LiZ09}. \pref{fig:weights} gives two examples of reference point distribution and the neighborhood of a reference point in the two- and three-objective cases.
\begin{figure}[htbp]
\centering
\subfloat[2-D case.]{\includegraphics[width=.4\linewidth]{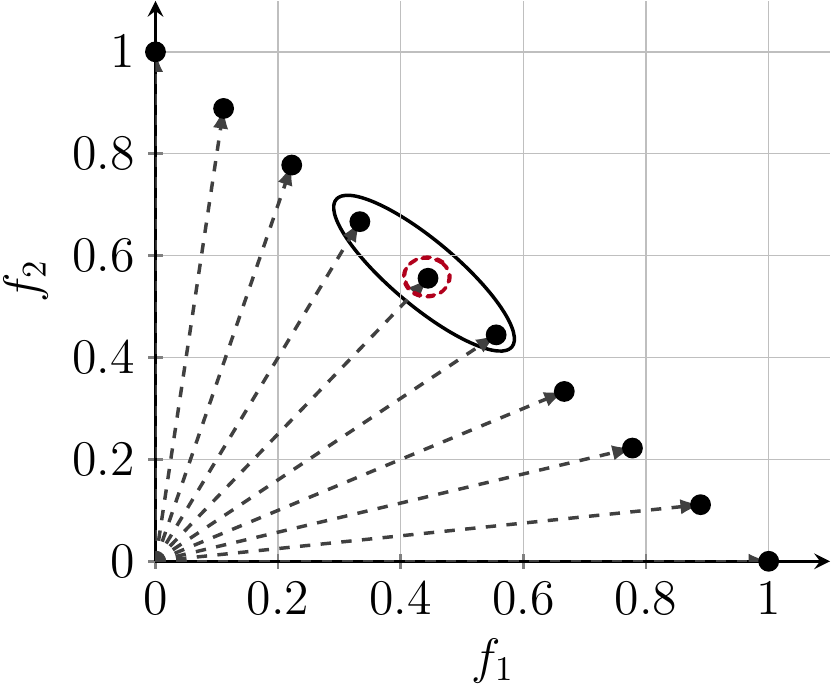}}
\subfloat[3-D case.]{\includegraphics[width=.4\linewidth]{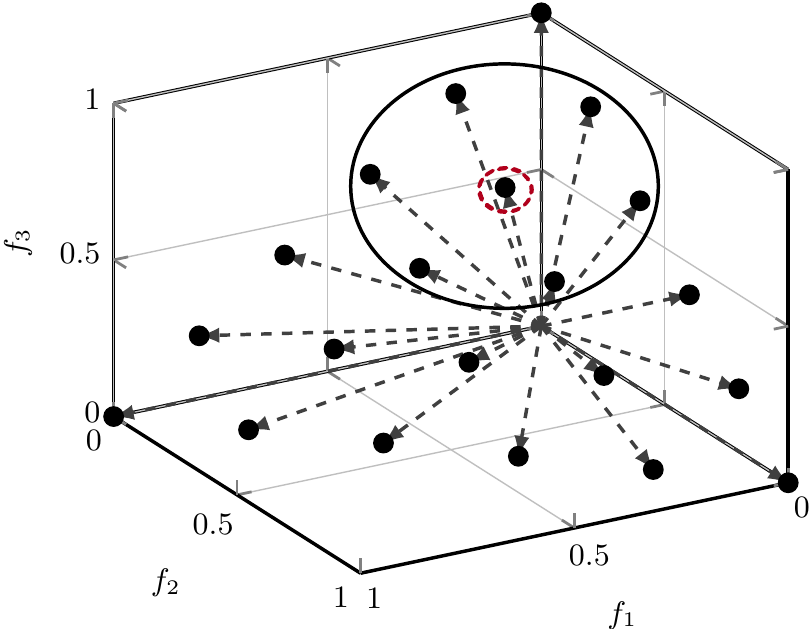}}
\caption{Illustration of reference points generated by the Das and Dennis' method~\cite{NBI}. The black circle represents the neighborhood of a particular reference point.}
\label{fig:weights}
\end{figure}

    \item In Step 2.1, to improve the exploration ability, there is a small probability $\delta=0.1$ to select mating parents from the whole population as suggested in~\cite{LiZ09}.
    \item In Step 2.3, $\mathbf{x}^c$ can update a particular reference point $\mathbf{w}$ if and only if $g(\mathbf{x}^c|\mathbf{w},\mathbf{z}^\ast)<g(\mathbf{x}|\mathbf{w},\mathbf{z}^\ast)$, where $\mathbf{x}$ is the solution originally associated with $\mathbf{w}$.
    \item In Step 2.3, $\mathbf{x}^c$ also has a small probability $\delta=0.1$ to update a subproblem from $W$, rather than merely in $B(i)$.
\end{itemize}

\subsubsection{NSGA-III}
This is an extension of NSGA-II for handling many-objective optimization problems. The subproblem in NSGA-III is to optimize the local crowdedness of the region associated with its corresponding reference point. In particular, it replaces the crowding distance with a reference point based density estimation. The general working mechanisms of NSGA-III is given as follows.
\begin{enumerate}[Step 1:]
    \item Initialize a population of solutions $P:=\{\mathbf{x}^i\}_{i=1}^N$, a set of reference points $W:=\{\mathbf{w}^i\}_{i=1}^N$.
    \item Use crossover and mutation to generate a population of offspring $Q$.
    \item Use non-dominated sorting~\cite{DebAPM02} to divide $R:=P\bigcup Q$ into several non-domination fronts $F_1,F_2,\cdots$.
    \item Starting from $F_1$, solutions are stored in a temporary archive $\overline{P}$ till its size for the first time equals or exceeds $N$, where $\overline{P}:=\bigcup_{i=1}^{\ell}F_i$. In particular, $F_{\ell}$ is the last acceptable non-domination front. If the size of $\overline{P}$ equals $N$, then let $P:=\overline{P}$ and go to Step 7; otherwise go to Step 5.
    \item Let $P:=\bigcup_{i=1}^{\ell-1} F_i$. Associate each member of $F_\ell$ with its closest reference point.
    \item A randomly chosen solution in the least crowded reference point is added into $P$. This process iterates till the size of $P$ equals $N$.
    \item If the stopping criteria is met, then stop and output $P$. Otherwise, go to Step 2.
\end{enumerate}

We would like to make some remarks on some important ingredients of the above NSGA-III procedure.
\begin{itemize}
    \item In Step 5, the association of a solution with a reference point is according to the shortest perpendicular distance between this solution and the reference line, starting from the origin and passing through the corresponding reference point. \pref{fig:association} gives a simple illustration of association in a two-dimensional scenario.
\begin{figure}[htbp]
\centering
\includegraphics[width=.4\linewidth]{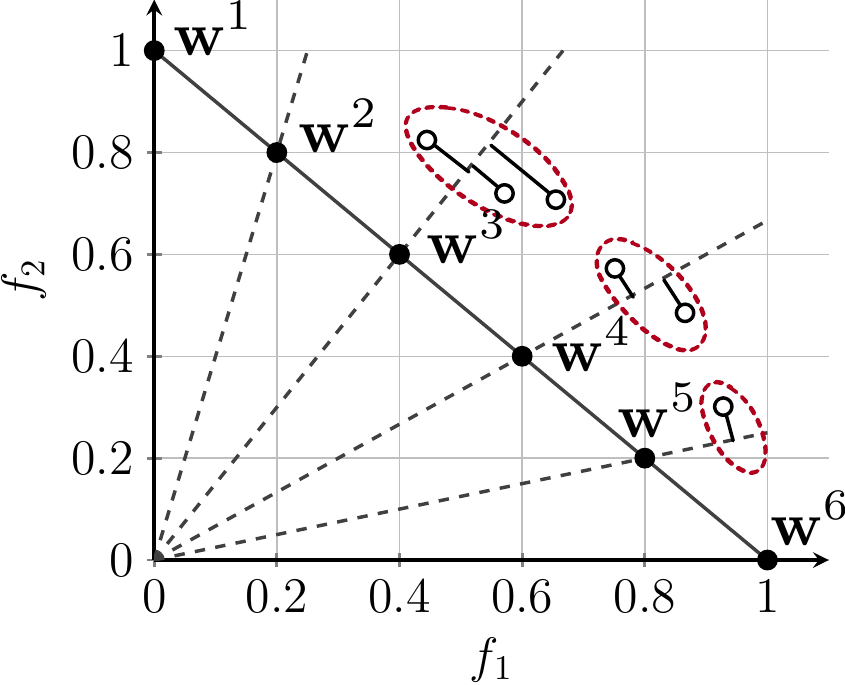}
\caption{Illustration of solution association of NSGA-III.}
\label{fig:association}
\end{figure}
    \item In Step 6, the crowdedness of a reference point is counted as the number of solutions associated with it. For example, as shown in~\pref{fig:association}, the crowdedness of $\mathbf{w}^3$ is 3. Note that the crowdedness information is updated after choosing a solution from $F_{\ell}$ and add it to $P$.
\end{itemize}

\subsection{Past Studies on Progressively Interactive Methods}
\label{sec:overview}

As mentioned in~\pref{sec:introduction}, there have been a plethora of studies to approximate DM's preferred solutions \textit{a priori}, \textit{posteriori} or \textit{interactively}. Since this paper mainly investigates the frequent involvements of a DM with an EMO algorithm, we do not intend to review \textit{a priori} and \textit{posteriori} approaches, except to encourage the interested readers to look at some recent survey papers~\cite{IEMO,BechikhKSG15,WangOJ17}.

Some recent studies periodically asked the DM to provide her/his preference information upon one or more pairs of alternative points found by an EMO algorithm. The information is then used to derive a value function that represents the DM's preference information. For example, Phelps and K\"oksalan~\cite{PhelpsK03} proposed an interactive EA that progressively constructs a linear value function, which is a weighted sum of objectives, by periodically asking the DM to rank pairs of solutions. Thereafter, the resulting value function is then used as the selection criteria of an EA to rank solutions. However, due to the use of a linear model, it might not be effective when the DM's \lq\lq golden\rq\rq\ value function is nonlinear. In~\cite{FowlerGKKMW10}, Fowler \textit{et al.} proposed to use convex preference cones to model the DM's preference information. In their developed interactive EMO algorithm, such cones are used to partially rank the population members and thus facilitate the fitness assignment. Instead of merely using a single value function, Jaszkiewicz~\cite{Jaszkiewicz07} proposed to use a set of linear value functions, each of which is a weighted sum of objectives, chosen from several randomly generated value functions to represent DM's preference information. Due to the use linear value function, it remains the same hallmark as~\cite{PhelpsK03} in handling nonlinear problems, especially when the DM's preferred solutions lie on a non-convex part of the PF.

In~\cite{DebSKW10}, Deb~\textit{et al.} developed a polynomial value function model that is expected to be useful for both linear and nonlinear problems. Specifically, to obtain the preference information, the DM is asked to rank a set of well distributed candidates periodically. Based on this order information, a polynomial value function model is fitted by solving a computationally intensive sequential quadratic programming procedure. Once a most discriminating value function has been identified, it is used to modify the Pareto dominance principle in NSGA-II in order to emphasize the reproduction and survival of preferred solutions. Moreover, the polynomial value function is also used to determine whether the overall optimization procedure should be terminated or not by performing a local search procedure.

In~\cite{BattitiP10}, Battiti and Passerini developed an interactive EMO algorithm that uses the support vector machine (SVM) for ranking~\cite{Joachims06} to represent complex value functions. Specifically, the DM is asked to rank (at least partially) some selected alternatives during the interaction session. This ranking information is then used to train a SVM, and the derived value function is used to replace the crowding distance in NSGA-II. Their empirical results suggested that the training of a SVM requires a relatively large number of solutions, whereas a small number of interactions seem to be sufficient to approximate the DM's golden value function.

In~\cite{BrankeGSZ09} and~\cite{BrankeGSZ15}, Branke~\textit{et al.} proposed an interactive EMO algorithm by ordinal regression which is able to build preference models compatible with preference information from holistic comparisons of solutions. During the interaction session, the DM is asked to rank a single pair of solutions. This information is used to update the additive value function model that is used in subsequent generations to rank incomparable solutions in terms of the Pareto dominance principle.

In~\cite{KorhonenMW86}, Korhonen~\textit{et al.} developed an interactive MO algorithm that progressively learns the DM's preference information by asking the DM to make a set of binary comparisons among several solutions. Specifically, a class of value functions are identified by solving a linear programming problem upon the preference information obtained from the interaction. In particular, they considered three classes of value functions, i.e., linear, quasi-concave and no pre-assumed forms. Based on this classification, they defined a dominance structure and determined the expected probabilities of finding new and better solutions either by search or choosing from several samples. Note that this algorithm terminates if the probability of finding better solutions is low and thus just outputs the currently found most preferred solution. As an extension, \cite{KorhonenMSW93} developed a sampling-based method to calculate the expected probabilities of finding better solutions.

%% file: proposal.tex

\section{Proposed Method}
\label{sec:proposal}

\begin{figure}[htbp]
\centering
\includegraphics[width=.5\linewidth]{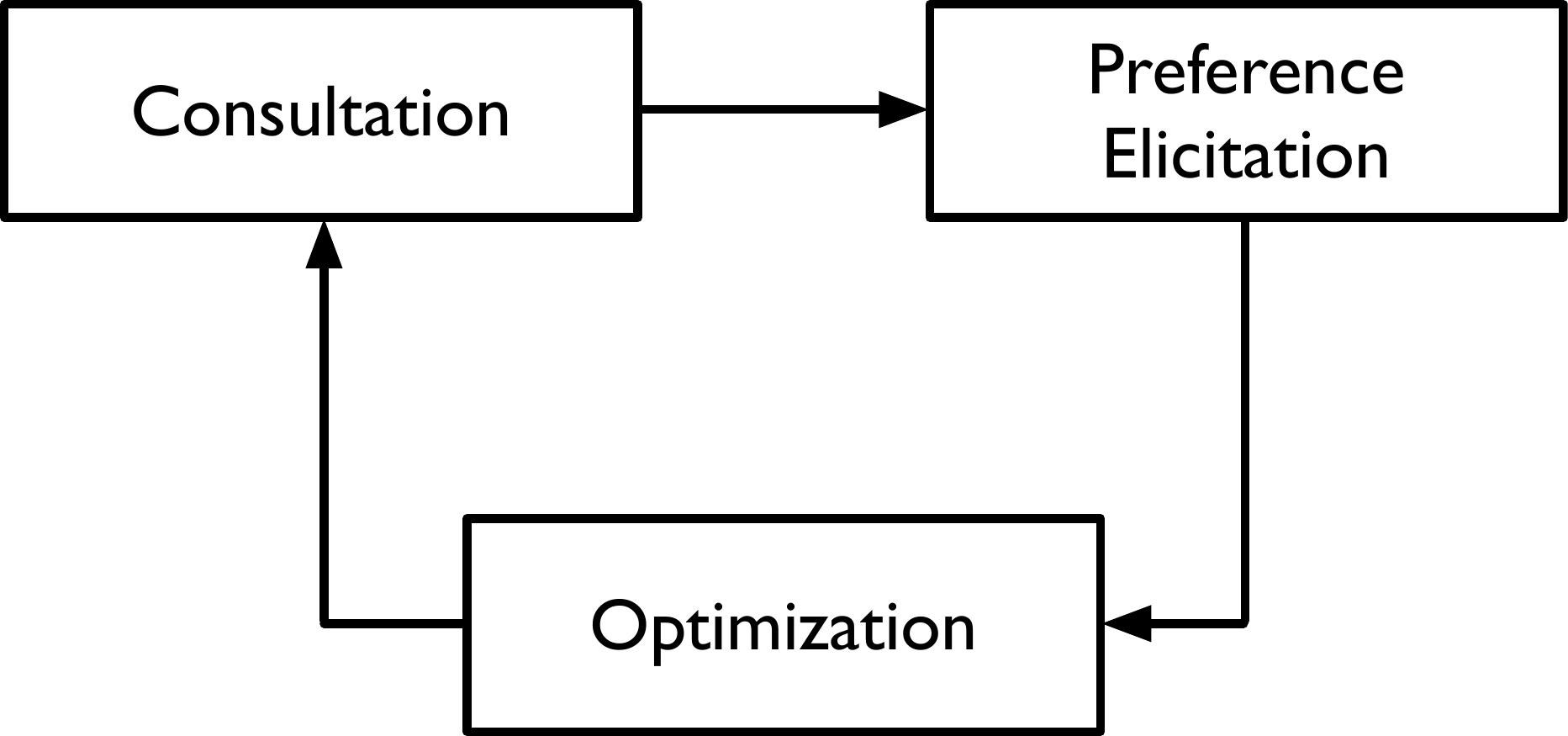}
\caption{Flowchart of the interactive framework.}
\label{fig:flowchart}
\end{figure}

As shown in~\pref{fig:flowchart}, our proposed interactive framework consists of three interdependent modules: consultation, preference elicitation and optimization. In principle, the optimization module can be any decomposition-based EMO algorithm. It uses the preference information provided by the preference elicitation module to find the DM's preferred solutions. In addition, it periodically supplies the consultation module with a few incumbent candidates for scoring. The consultation module is the interface by which the DM interacts with the optimization procedure. It progressively learns an AVF, which represents the DM's preference information, from the DM's behavior. The preference elicitation module translates the preference information, learned from the consultation module, into the form that can be used in the optimization module. In the following paragraphs, we will introduce the technical details of each module step by step. 

\subsection{Consultation Module}
\label{sec:consultation}

The consultation module is the interface where the DM interacts with, and expresses her/his preference information to the optimization module. In principle, there are various ways to represent the DM's preference information. In this paper, we assume that the DM's preference information is represented as a value function. It assigns a solution a score that represents its desirability to the DM. The consultation module mainly aims to progressively learn an AVF that approximates the DM's golden value function, which is unknown \textit{a priori}, by asking the DM to score a few incumbent candidates. We argue that it is labor-intensive to consult the DM every generation. Furthermore, as discussed in~\cite{BattitiP10}, consulting the DM at the early stage of the evolution might be detrimental to the decision-making procedure, since the DM can hardly make a reasonable judgement on poorly converged solutions. In this paper, we fix the number of consultations. Before the first consultation session, the EMO algorithm runs as usual without considering any DM's preference information. Afterwards, the consultation session happens every $\tau>1$ generations.

To approximate the DM's preference information, we need to address two major questions: 1) which solutions can be used for scoring? and 2) how to learn an appropriate AVF? 

\subsubsection{Scoring}
\label{sec:label}

A na\"ive strategy is to ask the DM to score all solutions in a population. In this case, the search is completely driven by the DM. This obviously increases her/his cognitive load thus has a high risk to cause her/his fatigue. Instead, during each consultation session, we only ask the DM to score a limited number (say $1\leq\mu\ll N$) of incumbent candidates chosen from the current population.

If it is at the first consultation session, we first initialize another $\mu$ \lq\lq seed\rq\rq\ reference points, which can either be generated by the Das and Dennis' method~\cite{NBI} or chosen from the reference points initialized in the optimization module. Afterwards, for each of these seed reference points, we find the nearest neighbor from the reference points initialized in the optimization module. Then, the solutions associated with these selected reference points are used as the initial incumbent candidates. Otherwise, at the latter consultation sessions, we use the AVF learned from the last consultation session to score the current population. The $\mu$ solutions having the best AVF values are deemed as the ones that are satisfied by the DM most. Accordingly, these $\mu$ solutions are used as the incumbent candidates. 

\subsubsection{Learning}
\label{sec:training} 

In principle, many off-the-shelf machine learning algorithms can be used to learn the AVF. In this paper, we treat it as a regression problem and use the Radius Basis Function network (RBFN)~\cite{RBF} to serve this purpose. In particular, RBFN, a single-layer feedforward neural network, is easy to train and its performance is relatively insensitive to the increase of the dimensionality. The idea of using RBFN as an approximation function was first proposed by Hardy~\cite{Hardy71} to fit irregular topological data.

Let $\mathcal{D}=\{(\mathbf{F}(\mathbf{x}^i), \psi(\mathbf{x}^i))\}_{i=1}^M$ denote the dataset for training the RBFN. The objective values of a solution $\mathbf{x}^i$ are the inputs and its corresponding value function $\psi(\mathbf{x}^i)$ scored by the DM is the output. In particular, we accumulate every $\mu$ solutions scored by the DM to form $\mathcal{D}$. An RBFN is a real-valued function $\Phi:\mathbb{R}^m\rightarrow\mathbb{R}$. Various RBFs can be used as the activation function of the RBFN, such as Gaussian, splines and multiquadrics. In this paper, we consider the following Gaussian function:
\begin{equation}
    \varphi=\exp(-\frac{\|\mathbf{F}(\mathbf{x})-\mathbf{c}\|}{\sigma^2}),
\end{equation} 
where $\sigma>0$ is the width of the Gaussian function. Accordingly, the AVF can be calculated as:
\begin{equation}
    \Phi(\mathbf{x})=\omega^0+\sum_{i=1}^{\mathrm{NR}}\omega^i\exp(-\frac{\|\mathbf{F}(\mathbf{x})-\mathbf{c}^i\|}{\sigma^2}),
\end{equation}
where $\mathrm{NR}$ is the number of RBFs, each of which is associated with a different center $\mathbf{c}^i$, $i\in\{1,\cdots,\mathrm{NR}\}$. $\omega^i$ is the network coefficient, and $\omega^0$ is a bias term, which can be set to the mean of the training data or 0 for simplicity. In our experiment, we use the RBFN program $\mathsf{newrb}$ provided by the Neural Network Toolbox from the MATLAB\footnote{\url{https://uk.mathworks.com/help/nnet/ug/radial-basis-neural-networks.html}}.

\subsection{Preference Elicitation Module}
\label{sec:preference}

As introduced in~\pref{sec:DEMO}, the decomposition-based EMO algorithm is originally designed to use a set of evenly distributed reference points $W=\{\mathbf{w}^i\}_{i=1}^N$ to approximate the whole PF. When considering the DM's preference information, the ROI becomes a partial region of the PF. A natural idea, which translates the DM's preference information into the form that can be used in a decomposition-based EMO algorithm, is to adjust the distribution of reference points. Specifically, the preference elicitation module uses the following four-step process to achieve this purpose.

\begin{enumerate}[Step 1:]
    \item Use $\Phi(\mathbf{x})$ learned in the consultation module to score each member of the current population $P$.
    \item Rank the population according to the scores assigned in Step 1, and find the top $\mu$ solutions. Reference points associated with these solutions are deemed as the promising ones, and store them in a temporary archive $W^U:=\{\mathbf{w}^{Ui}\}_{i=1}^{\mu'}$.
    \item For $i=1$ to $\mu'$ do 
        \begin{enumerate}[Step 3.1:]
            \item If $\Phi(\mathbf{x}^{Ui})<g(\mathbf{x}^{best}|\mathbf{w}^{best},\mathbf{z}^\ast)$, then go to Step 3.2. Otherwise, move the remaining reference points toward $\mathbf{w}^{best}$ as follows:
            \begin{equation}
                w_j=w_j+\eta\times(w^{best}_j-w_j),
                \label{eq:weight_update}
            \end{equation}
            where $j\in\{1,\cdots,m\}$. Terminate the for-loop and go to Step 4.
            \item Find the $\lceil\frac{N-\mu'}{\mu'}\rceil$ closest reference points to $\mathbf{w}^{Ui}$ according to their Euclidean distances.
            \item Move each of these reference points toward $\mathbf{w}^{Ui}$ according to~\pref{eq:weight_update}, where $\mathbf{w}^{best}$ is replaced with $\mathbf{w}^{Ui}$.
            \item Temporarily remove these reference points from $W$ and go to Step 3.
        \end{enumerate}
    \item Output the adjusted reference points as the new $W$.
\end{enumerate}

We would like to make some remarks on some important ingredients of the above process.

\begin{itemize}
    \item In Step 1, the score of a solution, evaluated by the AVF learned in the consultation module, indicates its satisfaction with respect to the DM's preference information.
    \item In the decomposition-based EMO algorithm, each solution should be associated with a reference point. Therefore, in Step 2, the rank of a solution indicates the importance of its associated reference point with respect to the DM's preference information. The reference points stored in $W^U$ are indexed according to the ranks of their associated solutions. In other words, $\mathbf{w}^{U1}$ represents the most important reference point, and so on.
    \item Furthermore, since a reference point might be associated with more than one solution (e.g., in NSGA-III), the number of promising reference points $\mu'$ might be smaller than $\mu$, i.e., $1\leq\mu'\leq\mu$.
    \item Step 3 is the main crux to adjust the distribution of reference points. The major purpose of this process is to move the other reference points toward those $\mu'$ promising ones. In particular, each of these promising reference points attracts around $\lceil\frac{N-\mu'}{\mu'}\rceil$ companions.
    \item In Step 3.1, $\mathbf{x}^{Ui}$ represents the solution associated with $\mathbf{w}^{Ui}$. $\mathbf{x}^{best}$ represents the best solution evaluated by the DM at the last consultation session, while $\mathbf{w}^{best}$ represents its associated reference point. The major purpose of Step 3.1 is to alleviate the risk of moving the reference points to a wrongly predicted promising one.
    \item In~\pref{eq:weight_update}, the step size $\eta$ controls the convergence rate toward the promising reference point. 
    \item Step 3.2 is similar to a clustering process, while we give the reference point, which has a higher rank, a higher priority to attract its companions.
\end{itemize}

\begin{figure}[htbp]
\centering
\subfloat[Original distribution.]{\includegraphics[width=.4\linewidth]{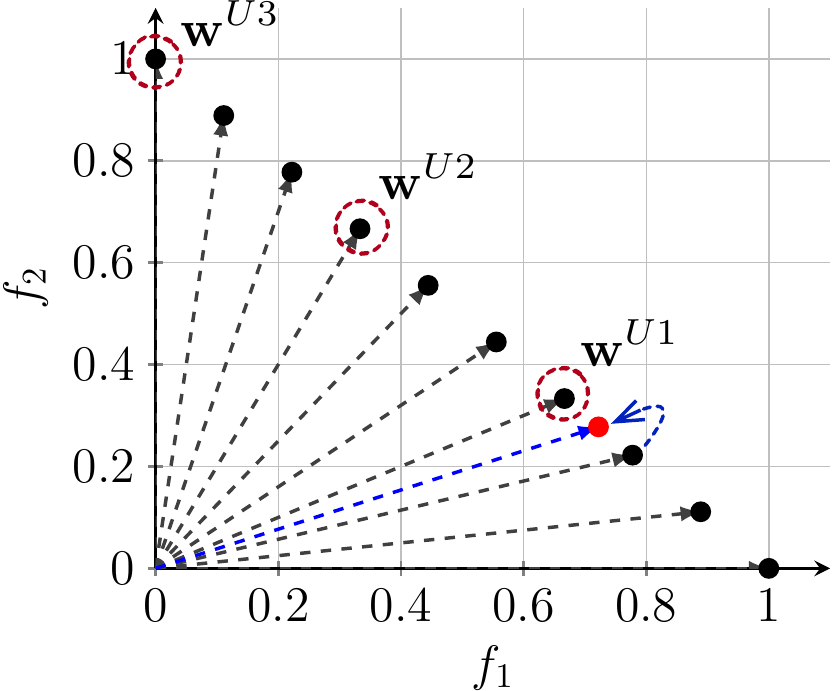}}
\subfloat[Adjusted distribution.]{\includegraphics[width=.4\linewidth]{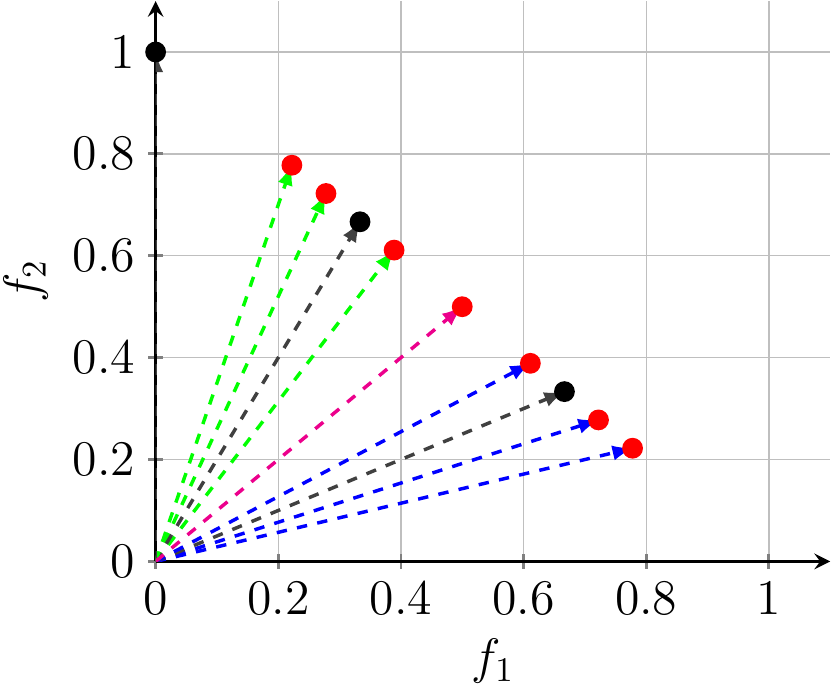}}
\caption{Illustration of the preference elicitation process.}
\label{fig:pref-elicit}
\end{figure}

To have a better understanding of this preference elicitation process, \pref{fig:pref-elicit} gives an example in a two-objective case. In particular, three promising reference points are highlighted by red circles. $\mathbf{w}^{U1}$ has the highest priority to attract its companions, and so on.

\subsection{Optimization Module}
\label{sec:optimization}

The optimization module is the search engine that progressively finds the DM's preferred solutions. In principle, any decomposition-based EMO algorithm can be used to serve this purpose. For the proof of principle purpose, this paper chooses MOEA/D and NSGA-III as the baseline algorithms, whose working mechanisms have been introduced in~\pref{sec:DEMO}. Note that MOEA/D and NSGA-III can be used in a plug-in manner without any modification except the reference points. In particular, the reference points used in MOEA/D and NSGA-III need to be adjusted by the preference elicitation module after every consultation session. As for the offspring reproduction, we use the popular simulated binary crossover (SBX)~\cite{SBX} and polynomial mutation~\cite{PolyMutation} for the proof of principle purpose.

%% file: settings.tex

\section{Experimental Settings}
\label{sec:settings}

To validate the effectiveness of our proposed interactive framework, we test the performance on benchmark problems with three to ten objectives. The interactive framework is embedded in MOEA/D and NSGA-III, and is respectively denoted as I-MOEA/D-PLVF and I-NSGA-III-PLVF. The widely used DTLZ~\cite{DebTLZ05} test problems are chosen to form the benchmark suite. Note that the DTLZ test problems are scalable to any number of objectives. Their formal definitions are described in Section I of the supplementary document.

The parameter settings of our proposed interactive framework are summarized as follows:
\begin{itemize}
\item number of incumbent candidates presented to the DM for scoring: $\mu=2m+1$ at the first consultation session and $\mu=10$ afterwards;
\item number of generations between two consecutive consultation sessions: $\tau=25$;
\item number of reference points and population size settings are given in~\pref{tab:popsize} as suggested in~\cite{LiDZK15};
\item number of function evaluations (FEs) is given in~\pref{tab:nFEs} as suggested in~\cite{LiDZK15}.
\item step size of the reference point update used in~\pref{eq:weight_update}: $\eta=0.5$;
\item crossover probability and the distribution index for the SBX operator: $p_c=1.0$ and $\eta_c=30$;
\item mutation probability and the distribution index for the polynomial mutation operator: $p_m=0.9$ and $\eta_m=20$;

\begin{table}[htbp]
\scriptsize
\caption{Number of reference points and population size.}
\centering
\begin{tabular}{|@{ }c@{ }|c|c|c|}
\hline
$m$ & $\sharp$ of reference points & I-NSGA-III-PLVF & I-MOEA/D-PLVF \\ \hline
3   & 91  & 92  & 91  \\\hline
5   & 210 & 212 & 210 \\\hline
8   & 156 & 156 & 156 \\\hline
10  & 275 & 276 & 275 \\\hline
\end{tabular}
\label{tab:popsize}
\end{table}

\begin{table}[htbp]
\scriptsize
\caption{Number of FEs for DTLZ test problems.}
\centering
\begin{tabular}{|c|c|c|c|c|}
\hline
Test instance & $m=3$ & $m=5$ & $m=8$ & $m=10$ \\\hline
DTLZ1 & 400 & 600 & 750 & 1,000 \\\hline
DTLZ2 & 250 & 350 & 500 & 750 \\\hline
DTLZ3 & 1,000 & 1,000 & 1,000 & 1,500 \\\hline
DTLZ4 & 600 & 1,000 & 1,250 & 2,000 \\\hline
\end{tabular}
\begin{tablenotes}
\item[1] Each cell only gives the number of generations. The corresponding number of FEs is each tuple times the corresponding population size of I-NSGA-III-PLVF as shown in~\pref{tab:popsize}.
\end{tablenotes} 
\label{tab:nFEs}
\end{table}

\end{itemize}

As mentioned in~\cite{BrankeGSZ15}, the empirical comparison of interactive EMO methods is tricky since a model of the DM's behavior is required yet unfortunately sophisticated to represent. In this paper, we use a pre-specified golden value function, which is unknown to an interactive EMO algorithm, to play as an artificial DM. Specifically, the DM is assumed to minimize the following nonlinear Tchebycheff function:
\begin{equation}
    \psi(\mathbf{x})=\max_{1\leq i\leq m}|f_i(\mathbf{x})-\mathbf{z}^{\ast}|/w^{\ast}_i,
    \label{eq:vf}
\end{equation}
where $\mathbf{z}^{\ast}$ is set to be the origin in our experiments, and $\mathbf{w}^{\ast}$ is the utopia weights that represents the DM's emphasis on different objectives. We consider two types of $\mathbf{w}^\ast$: one targets the preferred solution on the middle region of the PF while the other targets the preferred solution on one side of the PF, i.e., biased toward a particular extreme. Since a $m$-objective problem has $m$ extremes, there are $m$ different choices for setting the biased $\mathbf{w}^{\ast}$. In our experiments, we randomly choose one for the proof-of-principle study. Since the Tchebycheff function is used as the value function and the analytical forms of the test problems are known, we can use the method suggested in~\cite{LiDZK15} to find the corresponding Pareto-optimal solution (also known as the DM's \lq\lq golden\rq\rq\ point) with respect to the given $\mathbf{w}^{\ast}$. Detailed settings of $\mathbf{w}^\ast$ and the corresponding DM's golden point are given in Section II of the supplementary document.

To evaluate the performance of an interactive EMO algorithm for approximating the ROI, we consider using the approximation error of the obtained population $P$ with respect to the DM's golden point $\mathbf{z}^r$ as the performance metric. Specifically, it is calculated as:
\begin{equation}
\mathbb{E}(P)=\min_{\mathbf{x}\in P}\mathsf{dist}(\mathbf{x},\mathbf{z}^r)
\end{equation}
where $\mathsf{dist}(\mathbf{x},\mathbf{z}^r)$ is the Euclidean distance between $\mathbf{z}^r$ and a solution $\mathbf{x}\in P$ in the objective space.

To demonstrate the importance of using the DM's preference information, we also compare I-MOEA/D-PLVF and I-NSGA-III-PLVF with their corresponding baseline algorithms without considering the DM's preference information. In our experiments, we run each algorithm independently 21 times with different random seeds. In the corresponding table, we show the results in terms of the median and the interquartile range (IQR) of the approximation errors obtained by different algorithms. To have a statistical sound comparison, we use the Wilcoxon signed-rank test with a 95\% confidence level to validate the significance of the better results.

%% file: experiments.tex

\section{Empirical Results}
\label{sec:results}

Our experiments are divided into three parts. First, we validate the effectiveness of our proposed interactive framework for finding the DM's preferred solution. Then, we empirically investigate the influence of the parameters associated with the interactive framework. At last, we investigate a scenario with random noises in the decision-making procedure.

\subsection{Performance Comparisons on DTLZ Test Problems}
\label{sec:DTLZ}

\begin{table*}[htbp]
\tiny
\centering
\caption{Performance comparisons of the approximation errors (median and the corresponding IQR) obtained by I-NSGA-III-PLVF and I-MOEA/D-PLVF versus their baseline MOEA/D and NSGA-III on DTLZ test problems.}
\label{tab:DTLZ-error}
\begin{tabular}{@{ }c@{ }@{ }c@{ }|c|c|c|c|c|c|c|c|}
\cline{3-10}
                    &     & \multicolumn{2}{c|}{DTLZ1}            & \multicolumn{2}{c|}{DTLZ2} & \multicolumn{2}{c|}{DTLZ3}            & \multicolumn{2}{c|}{DTLZ4}             \\ \hline
\multicolumn{1}{|@{}c@{}|}{$m$}                 & ROI & I-MOEA/D-PLVF    & MOEA/D           & I-MOEA/D-PLVF     & MOEA/D           & I-MOEA/D-PLVF    & MOEA/D           & I-MOEA/D-PLVF     & MOEA/D           \\ \hline
\multicolumn{1}{|@{}c@{}|}{\multirow{2}{*}{3}}  & $c$ & \multicolumn{1}{>{\columncolor{mycyan}}c}{\textbf{0.00042(2.87E-3)}} & 0.03104(3.18E-3) & \multicolumn{1}{>{\columncolor{mycyan}}c}{\textbf{0.01026(1.78E-2)}} & 0.10300(6.35E-3) & \multicolumn{1}{>{\columncolor{mycyan}}c}{\textbf{0.00072(7.26E-3)}} & 0.10553(1.59E-3) & \multicolumn{1}{>{\columncolor{mycyan}}c}{\textbf{0.01302(2.78E-2)}} & 0.10421(1.89E-3) \\ \cline{2-10}
\multicolumn{1}{|@{}c@{}|}{}                    & $b$ & \multicolumn{1}{>{\columncolor{mycyan}}c}{\textbf{0.00147(2.87E-3)}} & 0.03103(3.30E-3) & \multicolumn{1}{>{\columncolor{mycyan}}c}{\textbf{0.00883(1.09E-2)}} & 0.09103(2.56E-3) & \multicolumn{1}{>{\columncolor{mycyan}}c}{\textbf{0.00281(1.09E-2)}} & 0.08678(7.75E-3) & \multicolumn{1}{>{\columncolor{mycyan}}c}{\textbf{0.00763(8.76E-3)}} & 0.09469(8.07E-3) \\ \hline
\multicolumn{1}{|@{}c@{}|}{\multirow{2}{*}{5}}  & $c$ & \multicolumn{1}{>{\columncolor{mycyan}}c}{\textbf{0.00417(1.73E-2)}} & 0.05262(1.90E-2) & \multicolumn{1}{>{\columncolor{mycyan}}c}{\textbf{0.01721(2.86E-2)}} & 0.24170(1.90E-2) & \multicolumn{1}{>{\columncolor{mycyan}}c}{\textbf{0.01128(8.77E-2)}} & 0.24420(4.62E-2) & \multicolumn{1}{>{\columncolor{mycyan}}c}{\textbf{0.02762(5.74E-2)}} & 0.25693(2.37E-3) \\ \cline{2-10}
\multicolumn{1}{|@{}c@{}|}{}                    & $b$ & \multicolumn{1}{>{\columncolor{mycyan}}c}{\textbf{0.01082(2.09E-2)}} & 0.07648(1.65E-2) & \multicolumn{1}{>{\columncolor{mycyan}}c}{\textbf{0.05082(4.73E-2)}} & 0.20491(1.45E-2) & \multicolumn{1}{>{\columncolor{mycyan}}c}{\textbf{0.01792(1.53E-1)}} & 0.21623(2.35E-2) & \multicolumn{1}{>{\columncolor{mycyan}}c}{\textbf{0.03716(6.28E-2)}} & 0.21200(6.66E-3) \\ \hline
\multicolumn{1}{|@{}c@{}|}{\multirow{2}{*}{8}}  & $c$ & \multicolumn{1}{>{\columncolor{mycyan}}c}{\textbf{0.00213(1.71E-2)}} & 0.01484(2.21E-3) & \multicolumn{1}{>{\columncolor{mycyan}}c}{\textbf{0.01625(1.79E-1)}} & 0.26152(1.52E-2) & \multicolumn{1}{>{\columncolor{mycyan}}c}{\textbf{0.06821(2.78E-1)}} & 0.42766(9.56E-3) & \multicolumn{1}{>{\columncolor{mycyan}}c}{\textbf{0.06538(8.62E-2)}} & 0.72358(1.07E-2) \\ \cline{2-10}
\multicolumn{1}{|@{}c@{}|}{}                    & $b$ & \multicolumn{1}{>{\columncolor{mycyan}}c}{\textbf{0.01012(1.03E-1)}} & 0.05534(1.12E-2) & \multicolumn{1}{>{\columncolor{mycyan}}c}{\textbf{0.04184(1.10E-1)}} & 0.12504(1.05E-2) & \multicolumn{1}{>{\columncolor{mycyan}}c}{\textbf{0.08697(1.63E-1)}} & 0.15739(1.32E-2) & \multicolumn{1}{>{\columncolor{mycyan}}c}{\textbf{0.12708(1.86E-1)}} & 0.21640(1.69E-2) \\ \hline
\multicolumn{1}{|@{}c@{}|}{\multirow{2}{*}{10}} & $c$ & \multicolumn{1}{>{\columncolor{mycyan}}c}{\textbf{0.12690(2.71E-1)}} & 0.17885(1.10E-3) & \multicolumn{1}{>{\columncolor{mycyan}}c}{\textbf{0.10871(1.62E-1)}} & 0.73855(8.54E-2) & \multicolumn{1}{>{\columncolor{mycyan}}c}{\textbf{0.21682(5.71E-1)}} & 0.73645(2.81E-2) & \multicolumn{1}{>{\columncolor{mycyan}}c}{\textbf{0.19273(2.63E-1)}} & 0.86756(1.07E-1) \\ \cline{2-10}
\multicolumn{1}{|c|}{}                          & $b$ & \multicolumn{1}{>{\columncolor{mycyan}}c}{\textbf{0.15428(1.77E-1)}} & 0.26343(5.05E-3) & \multicolumn{1}{>{\columncolor{mycyan}}c}{\textbf{0.11829(2.08E-1)}} & 0.25957(2.88E-2) & \multicolumn{1}{>{\columncolor{mycyan}}c}{\textbf{0.16287(2.55E-1)}} & 0.33443(6.99E-2) & \multicolumn{1}{>{\columncolor{mycyan}}c}{\textbf{0.10175(3.28E-1)}} & 0.20545(4.21E-2) \\ \hline\hline
\multicolumn{1}{|@{}c@{}|}{$m$}                 & ROI & I-NSGA-III-PLVF   & NSGA-III          & I-NSGA-III-PLVF    & NSGA-III & I-NSGA-III-PLVF  & NSGA-III          & I-NSGA-III-PLVF    & NSGA-III          \\ \hline
\multicolumn{1}{|@{}c@{}|}{\multirow{2}{*}{3}}  & $c$ & \multicolumn{1}{>{\columncolor{mycyan}}c}{\textbf{0.00027(7.10E-4)}}  & 0.10382(3.56E-3) & \multicolumn{1}{>{\columncolor{mycyan}}c}{\textbf{0.00033(4.12E-4)}}  & 0.13951(5.20E-2) & \multicolumn{1}{>{\columncolor{mycyan}}c}{\textbf{0.00273(2.32E-4)}} & 0.14149(2.58E-2) & \multicolumn{1}{>{\columncolor{mycyan}}c}{\textbf{0.00518(2.19E-4)}}  & 0.13346(8.21E-2) \\ \cline{2-10}
\multicolumn{1}{|@{}c@{}|}{}                    & $b$ & \multicolumn{1}{>{\columncolor{mycyan}}c}{\textbf{0.00077(3.10E-4)}}  & 0.02822(2.42E-3) & \multicolumn{1}{>{\columncolor{mycyan}}c}{\textbf{0.00067(1.68E-4)}}  & 0.03979(2.18E-3) & \multicolumn{1}{>{\columncolor{mycyan}}c}{\textbf{0.00077(9.23E-5)}} & 0.04594(1.02E-2) & \multicolumn{1}{>{\columncolor{mycyan}}c}{\textbf{0.00748(6.69E-5)}}  & 0.03752(4.26E-3) \\ \hline
\multicolumn{1}{|@{}c@{}|}{\multirow{2}{*}{5}}  & $c$ & \multicolumn{1}{>{\columncolor{mycyan}}c}{\textbf{0.03127(4.12E-2)}}  & 0.21898(4.83E-3) & \multicolumn{1}{>{\columncolor{mycyan}}c}{\textbf{0.00536(2.76E-2)}}  & 0.24637(2.83E-2) & \multicolumn{1}{>{\columncolor{mycyan}}c}{\textbf{0.03192(2.48E-2)}} & 0.27167(3.13E-2) & \multicolumn{1}{>{\columncolor{mycyan}}c}{\textbf{0.01721(2.05E-2)}}  & 0.26652(1.93E-2) \\ \cline{2-10}
\multicolumn{1}{|@{}c@{}|}{}                    & $b$ & \multicolumn{1}{>{\columncolor{mycyan}}c}{\textbf{0.06781(3.21E-2)}}  & 0.20088(2.18E-2) & \multicolumn{1}{>{\columncolor{mycyan}}c}{\textbf{0.00142(1.79E-2)}}  & 0.21689(3.86E-2) & \multicolumn{1}{>{\columncolor{mycyan}}c}{\textbf{0.07622(1.47E-2)}} & 0.20411(1.84E-2) & \multicolumn{1}{>{\columncolor{mycyan}}c}{\textbf{0.03721(8.56E-3)}}  & 0.21169(1.13E-2) \\ \hline
\multicolumn{1}{|@{}c@{}|}{\multirow{2}{*}{8}}  & $c$ & \multicolumn{1}{>{\columncolor{mycyan}}c}{\textbf{0.08332(4.29E-2)}}  & 0.17453(5.29E-2) & \multicolumn{1}{>{\columncolor{mycyan}}c}{\textbf{0.06562(3.11E-2)}}  & 0.30752(1.27E-1) & \multicolumn{1}{>{\columncolor{mycyan}}c}{\textbf{0.07570(4.34E-2)}} & 0.21631(1.74E-1) & \multicolumn{1}{>{\columncolor{mycyan}}c}{\textbf{0.08773(2.88E-3)}}  & 0.20743(9.74E-2) \\ \cline{2-10}
\multicolumn{1}{|@{}c@{}|}{}                    & $b$ & \multicolumn{1}{>{\columncolor{mycyan}}c}{\textbf{0.07793(1.02E-1)}}  & 0.18973(2.62E-2) & \multicolumn{1}{>{\columncolor{mycyan}}c}{\textbf{0.05571(1.53E-2)}}  & 0.21554(1.82E-2) & \multicolumn{1}{>{\columncolor{mycyan}}c}{\textbf{0.12148(7.82E-2)}} & 0.22883(3.18E-2) & \multicolumn{1}{>{\columncolor{mycyan}}c}{\textbf{0.07242(7.62E-2)}}  & 0.28123(2.49E-2) \\ \hline
\multicolumn{1}{|@{}c@{}|}{\multirow{2}{*}{10}} & $c$ & \multicolumn{1}{>{\columncolor{mycyan}}c}{\textbf{0.11731(2.11E-1)}}  & 0.22708(1.29E-2) & \multicolumn{1}{>{\columncolor{mycyan}}c}{\textbf{0.07922(3.03E-1)}}  & 0.87988(1.02E-2) & \multicolumn{1}{>{\columncolor{mycyan}}c}{\textbf{0.18728(3.77E-1)}} & 0.88635(4.28E-3) & \multicolumn{1}{>{\columncolor{mycyan}}c}{\textbf{0.13752(3.20E-1)}}  & 0.87598(1.82E-2) \\ \cline{2-10}
\multicolumn{1}{|c|}{}                          & $b$ & \multicolumn{1}{>{\columncolor{mycyan}}c}{\textbf{0.12712(1.71E-1)}}  & 0.16275(3.28E-2) & \multicolumn{1}{>{\columncolor{mycyan}}c}{\textbf{0.11082(1.88E-1)}}  & 0.23098(1.28E-3) & \multicolumn{1}{>{\columncolor{mycyan}}c}{\textbf{0.21781(2.38E-1)}} & 0.24970(2.91E-2) & \multicolumn{1}{>{\columncolor{mycyan}}c}{\textbf{0.11683(2.02E-1)}}  & 0.22976(2.29E-2) \\ \hline
\end{tabular}

\begin{tablenotes}
\item[1] The ROI column gives the type of the DM supplied utopia weights. $c$ indicates the preference on the middle region of the PF while $b$ indicates the preference on an extreme. All better results are with statistical significance according to Wilcoxon signed-rank test with a 95\% confidence level, and are highlighted in bold face with a gray background.
\end{tablenotes}

\end{table*}

From the results shown in~\pref{tab:DTLZ-error}, we observe the overwhelming superiority of I-MOEA/D-PLVF and I-NSGA-III-PLVF, over the baseline MOEA/D and NSGA-III, for approximating the DM preferred solution. In particular, they obtain statistically significantly better metric values (i.e., smaller approximation error) on all test problems. In the following paragraphs, we discuss the results from three aspects.

\begin{itemize}
    \item \pref{fig:DTLZ_3Dc} to \pref{fig:DTLZ_10Db} plot the populations (with respect to the best approximation error) obtained by different algorithms. Note that since the observations on DTLZ3 and DTLZ4 test problems are similar to those on DTLZ2 test problem, we only show the plots on DTLZ1 and DTLZ2 test problems in this paper while the complete results are put in Section III of the supplementary document. From these plots, we can observe that both I-MOEA/D-PLVF and I-NSGA-III-PLVF are always able to find solutions that well approximate the unknown DM's golden point in a decent accuracy as shown in~\pref{tab:DTLZ-error}. In contrast, since the baseline MOEA/D and NSGA-III are designed to approximate the whole PF, it is not surprised to see that most of their solutions are away from the DM's golden point. Although some of the solutions obtained by the baseline MOEA/D and NSGA-III can by chance pass the ROI, i.e., the vicinity of the DM's golden point, they still have a observable distance from the DM's golden point. Moreover, the other solutions away from the ROI will unarguably result in the cognitive noise to \textit{posteriori} decision-making procedure, especially for problems that have many objectives.
    
\begin{figure}[htbp]
\centering
\includegraphics[width=\linewidth]{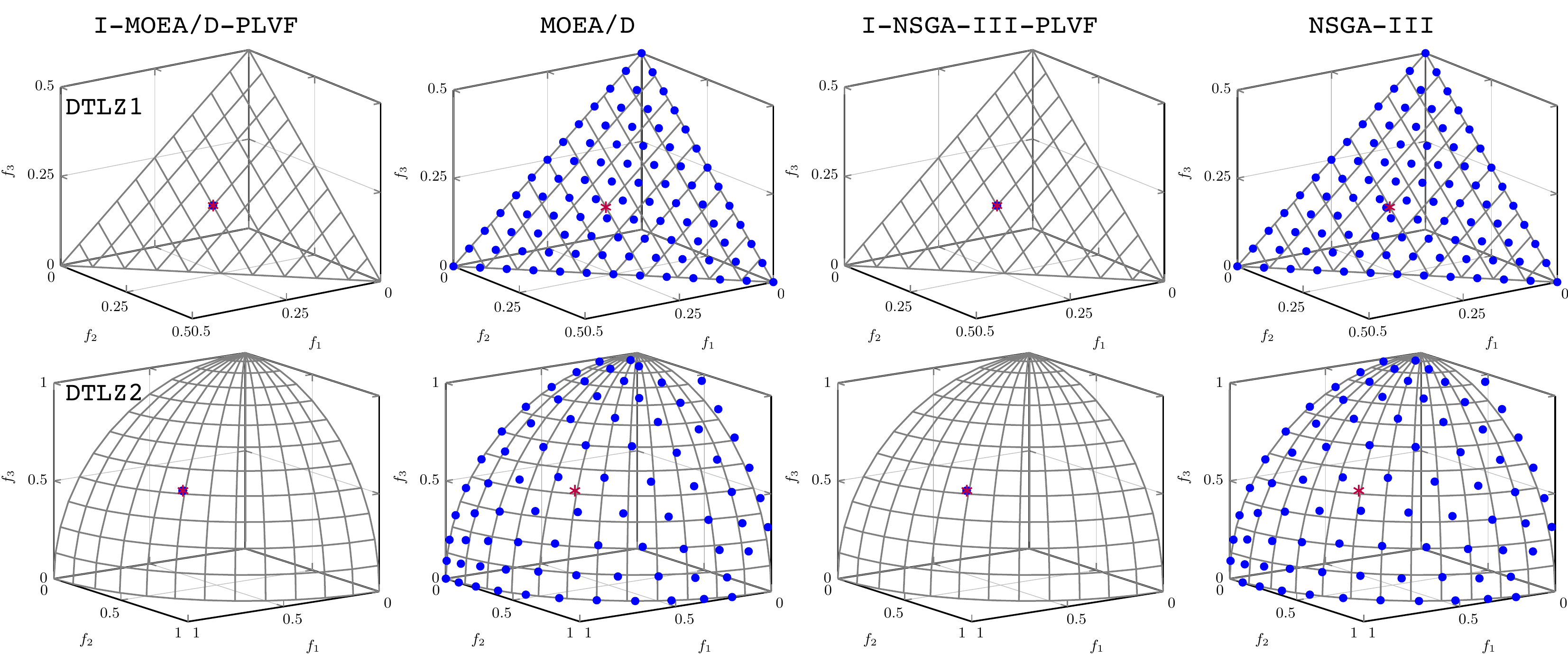}
\caption{Solutions obtained on 3-objective DTLZ1 and DTLZ2 test problems where $\mathbf{z}^r$, which prefers the middle region of the PF, is represented as the red asterisk.}
\label{fig:DTLZ_3Dc}
\end{figure}

\begin{figure}[htbp]
\centering
\includegraphics[width=\linewidth]{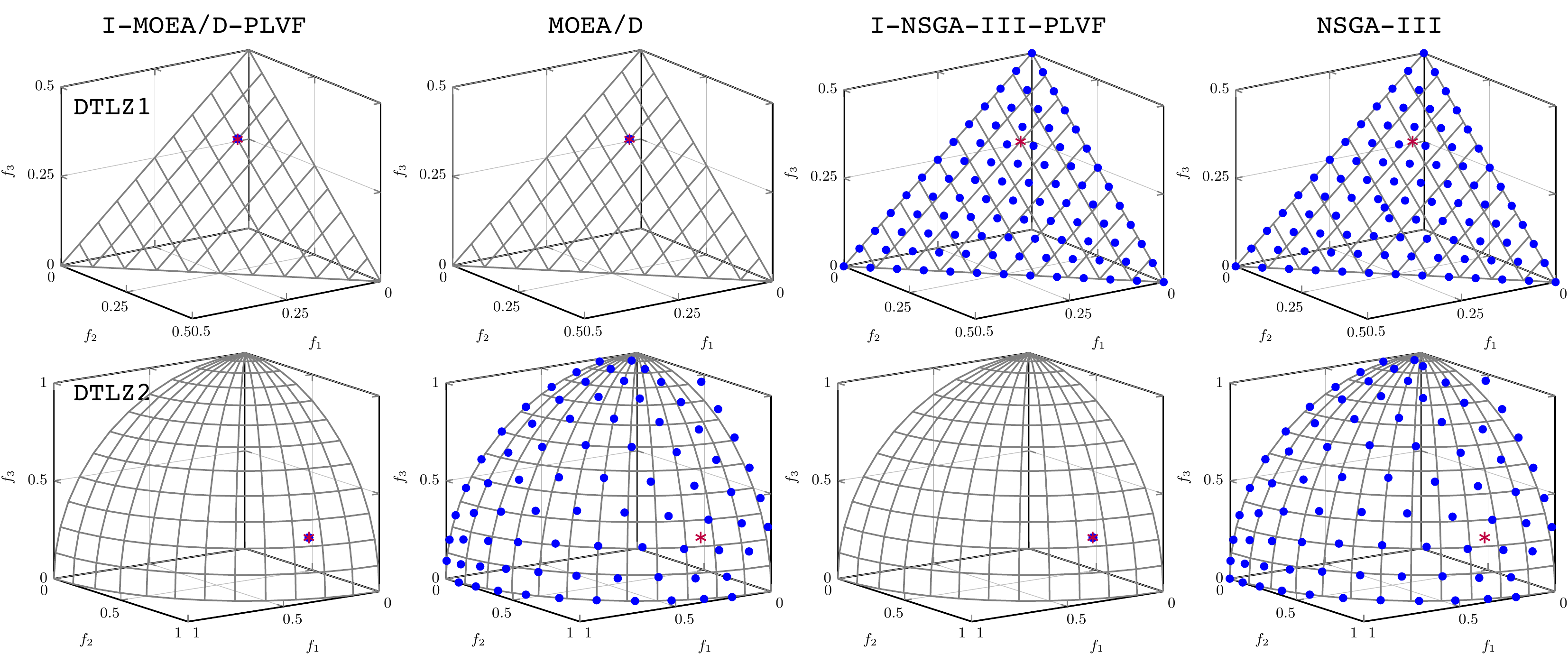}
\caption{Solutions obtained on 3-objective DTLZ1 and DTLZ2 test problems where $\mathbf{z}^r$, which prefers one side of the PF, is represented as the red asterisk.}
\label{fig:DTLZ_3Db}
\end{figure}

\begin{figure}[htbp]
\centering
\includegraphics[width=\linewidth]{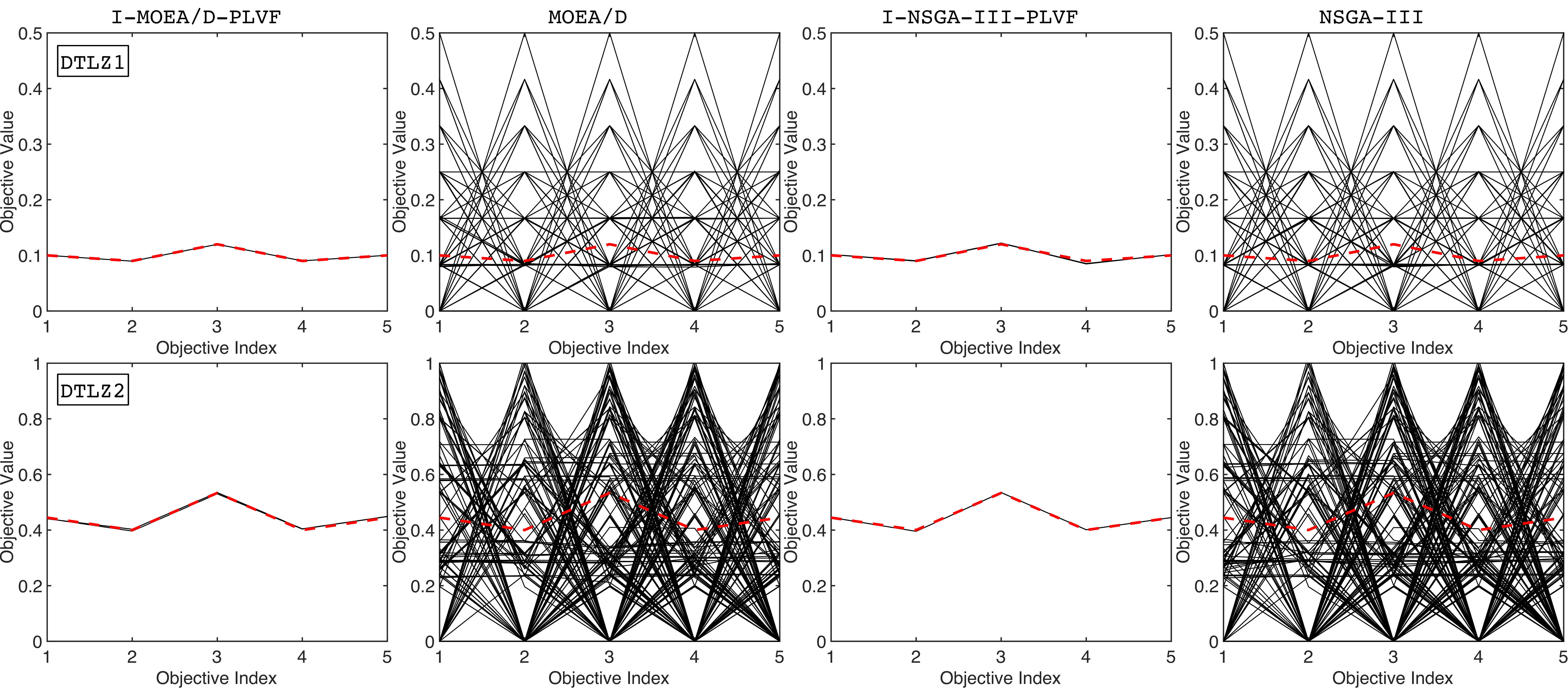}
\caption{Solutions obtained on 5-objective DTLZ1 and DTLZ2 test problems where $\mathbf{z}^r$, which prefers the middle region of the PF, is represented as the red dotted line.}
\label{fig:DTLZ_5Dc}
\end{figure}

\begin{figure}[htbp]
\centering
\includegraphics[width=\linewidth]{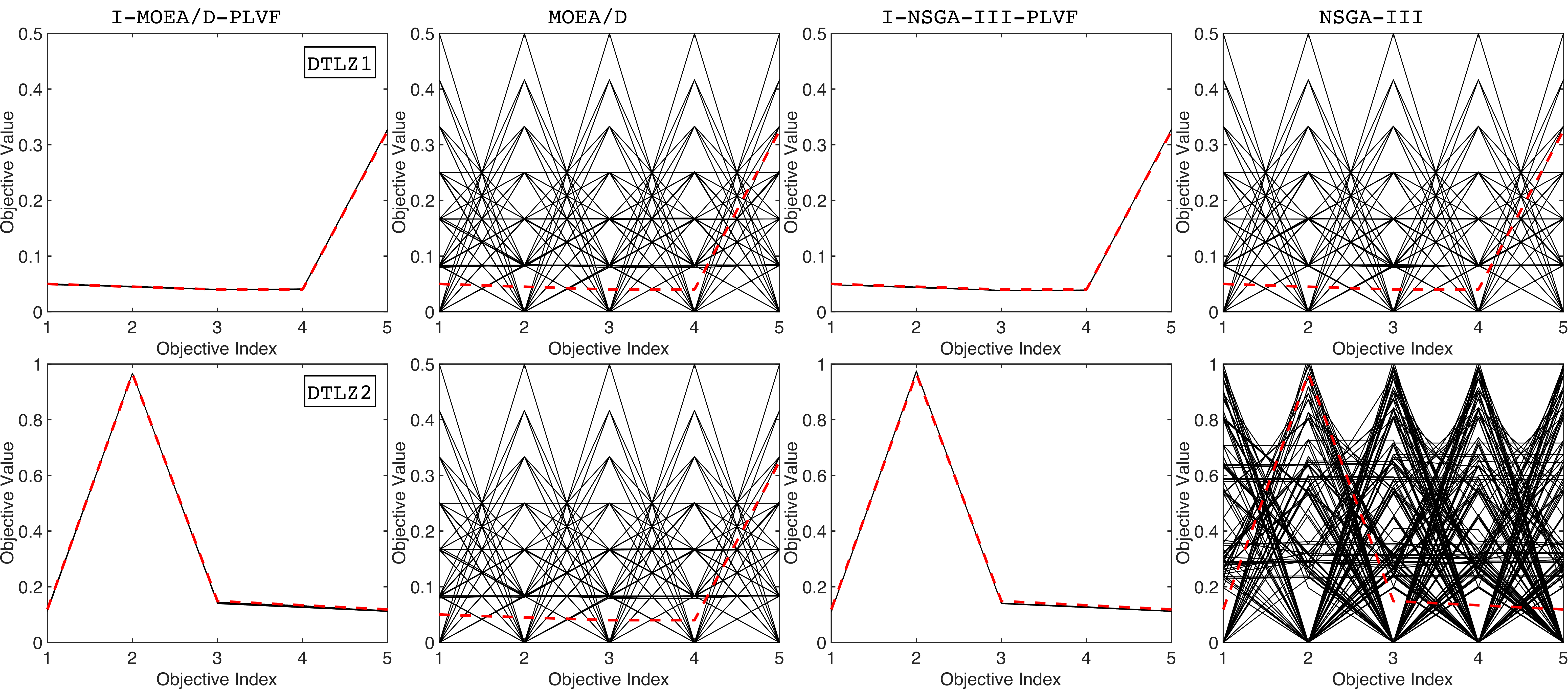}
\caption{Solutions obtained on 5-objective DTLZ1 and DTLZ2 test problems where $\mathbf{z}^r$, which prefers one side of the PF, is represented as the red dotted line.}
\label{fig:DTLZ_5Db}
\end{figure}

\begin{figure}[htbp]
\centering
\includegraphics[width=\linewidth]{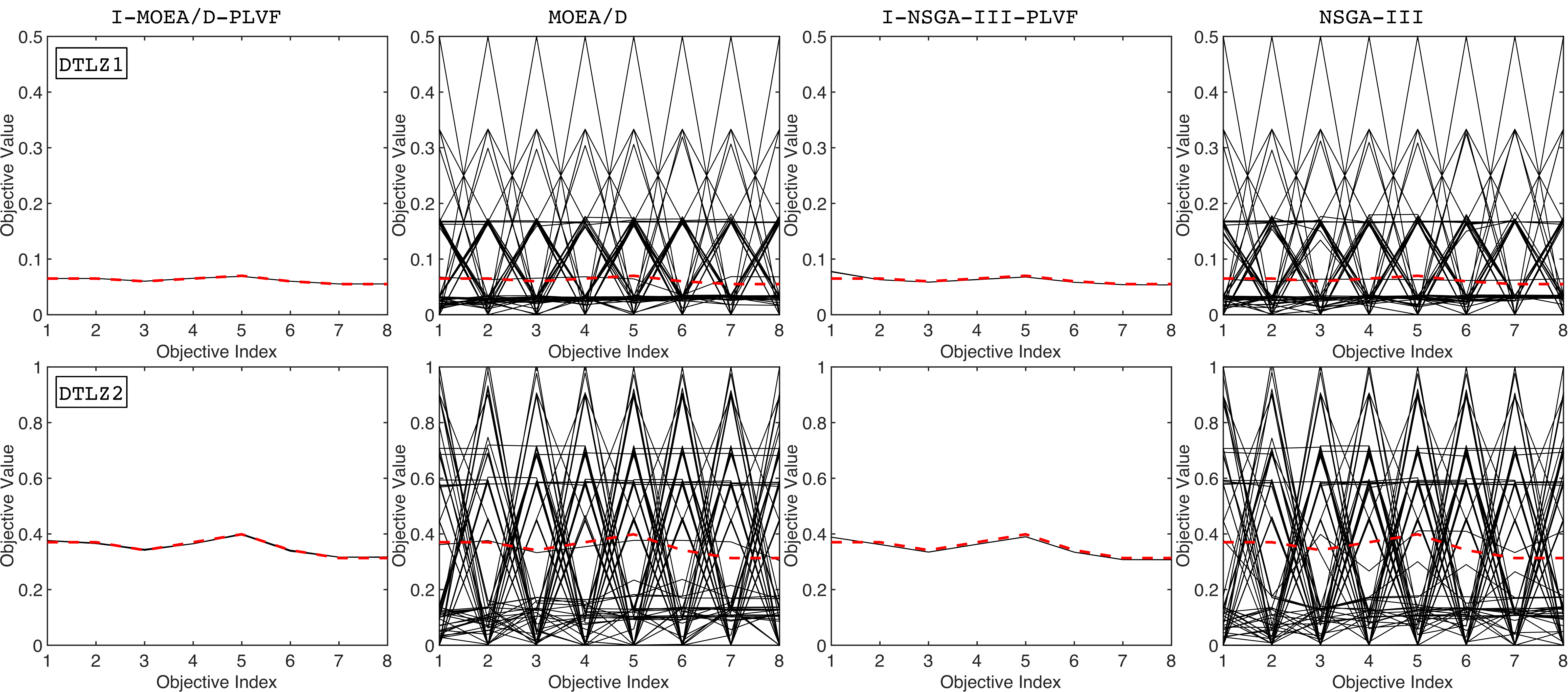}
\caption{Solutions obtained on 8-objective DTLZ1 and DTLZ2 test problems where $\mathbf{z}^r$, which prefers the middle region of the PF, is represented as the red dotted line.}
\label{fig:DTLZ_8Dc}
\end{figure}

\begin{figure}[htbp]
\centering
\includegraphics[width=\linewidth]{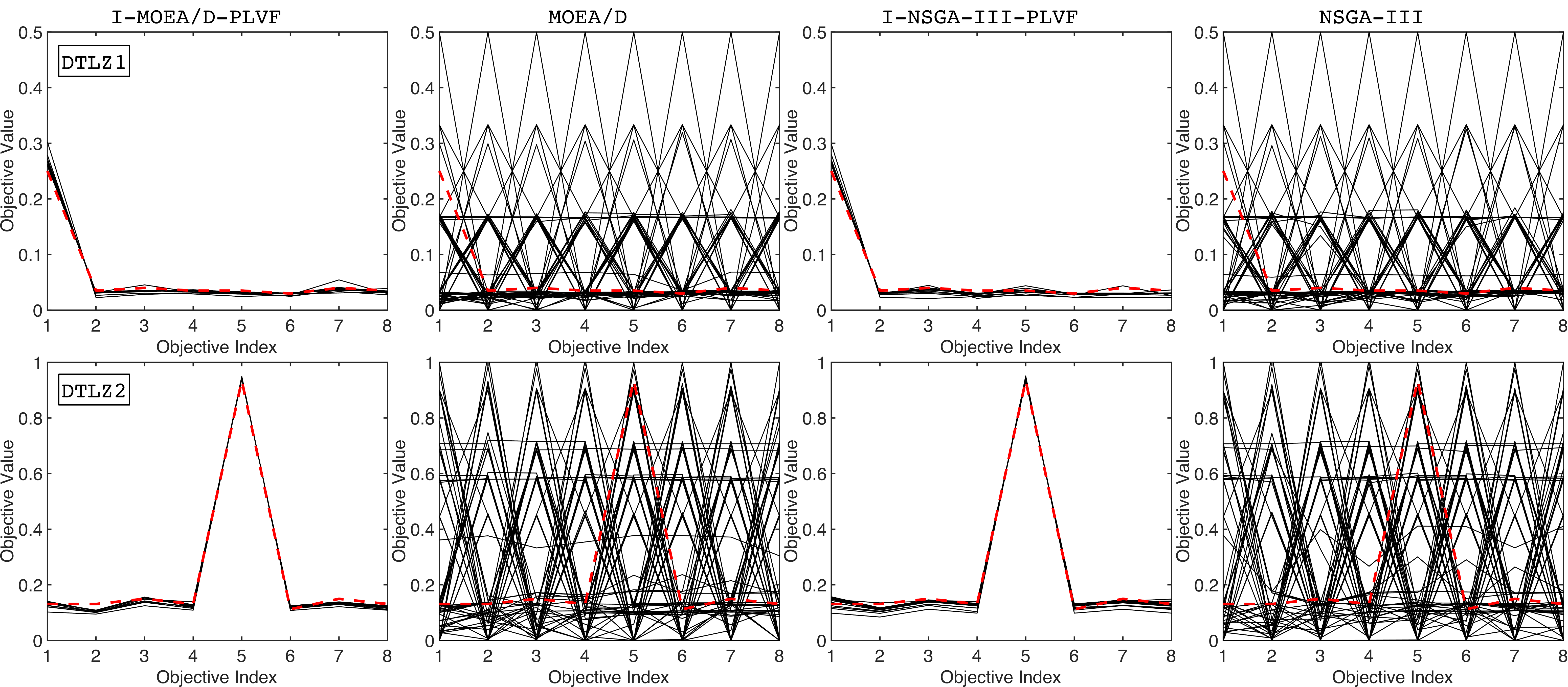}
\caption{Solutions obtained on 8-objective DTLZ1 and DTLZ2 test problems where $\mathbf{z}^r$, which prefers one side of the PF, is represented as the red dotted line.}
\label{fig:DTLZ_8Db}
\end{figure}

\begin{figure}[htbp]
\centering
\includegraphics[width=\linewidth]{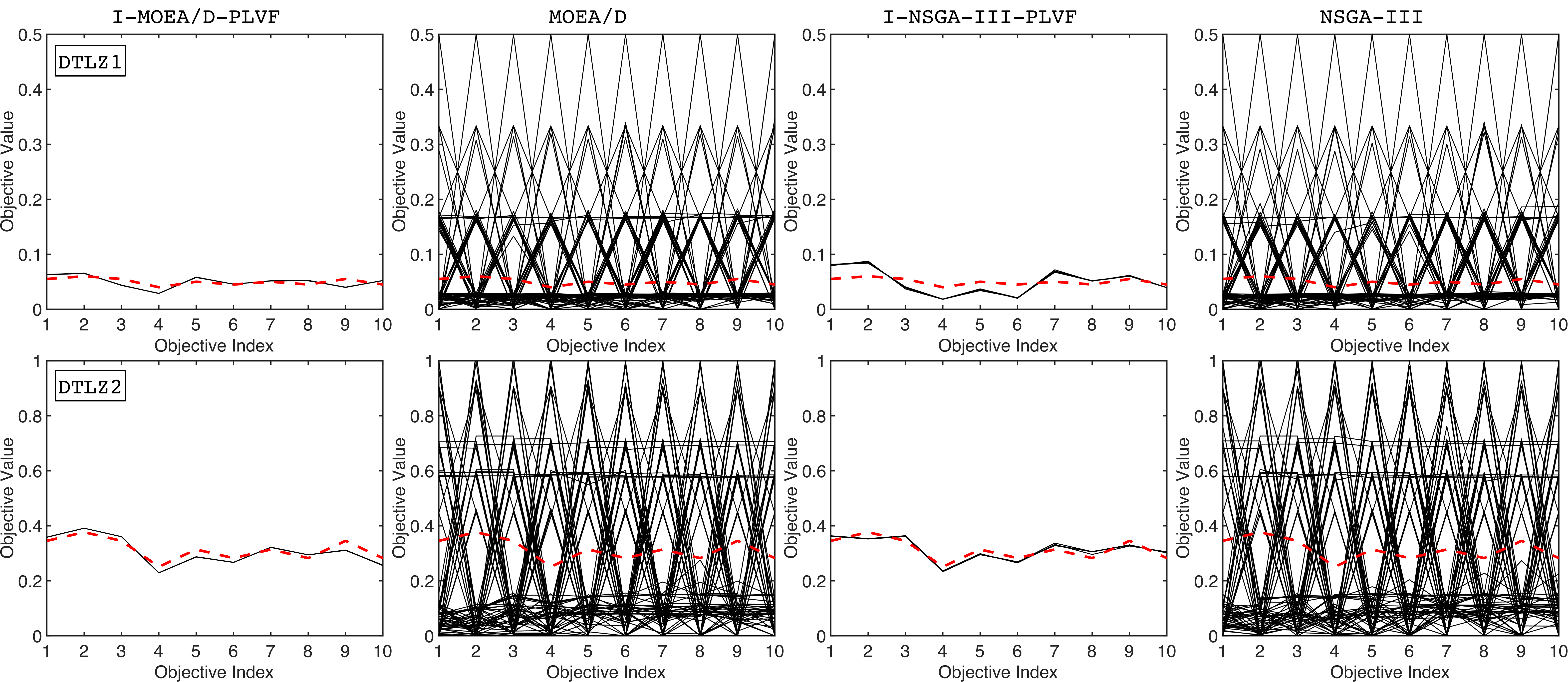}
\caption{Solutions obtained on 10-objective DTLZ1 and DTLZ2 test problems where $\mathbf{z}^r$, which prefers the middle region of the PF, is represented as the red dotted line.}
\label{fig:DTLZ_10Dc}
\end{figure}

\begin{figure}[htbp]
\centering
\includegraphics[width=\linewidth]{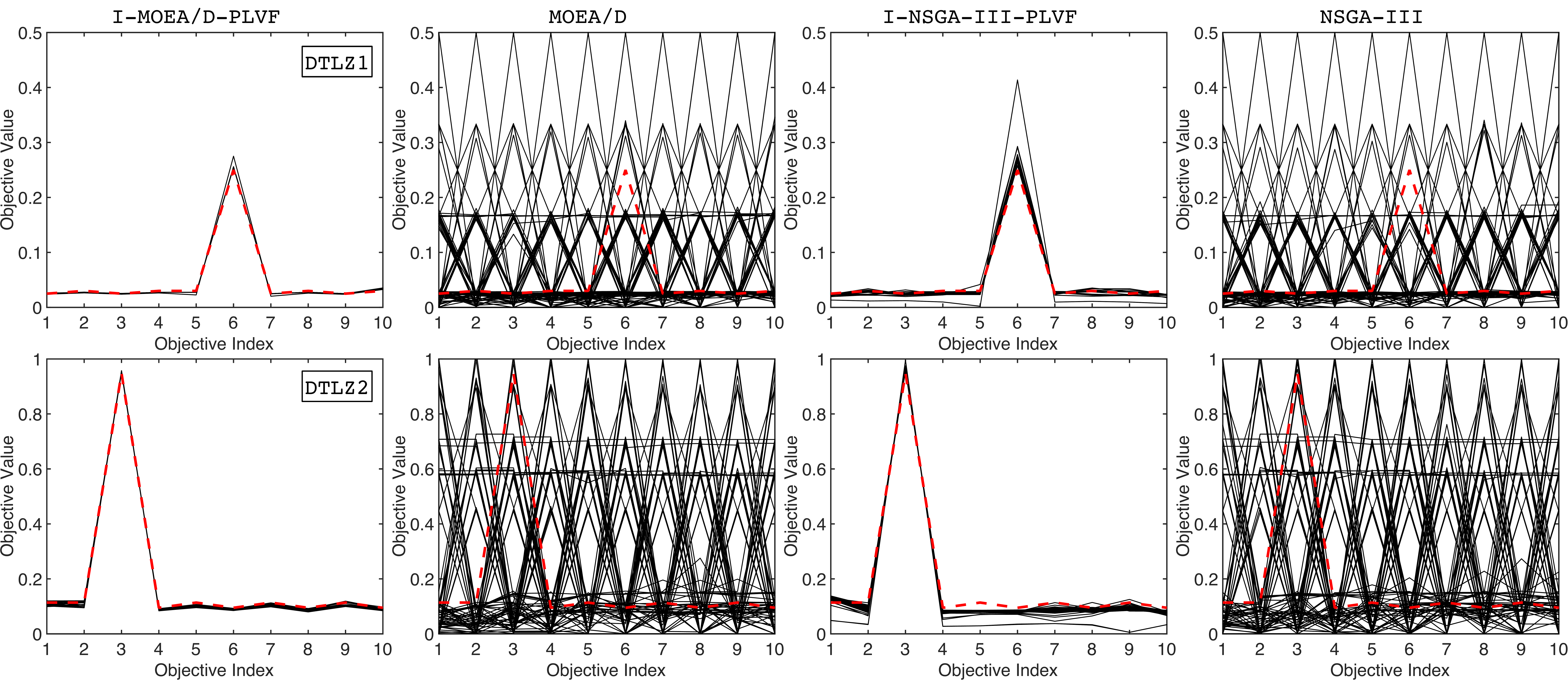}
\caption{Solutions obtained on 10-objective DTLZ1 and DTLZ2 test problems where $\mathbf{z}^r$, which prefers one side of the PF, is represented as the red dotted line.}
\label{fig:DTLZ_10Db}
\end{figure}

\item From the results shown in~\pref{tab:DTLZ-error}, we find that it seems to be more difficult for the baseline MOEA/D and NSGA-III to find the DM's preferred solution on the middle region of the PF than those biased toward a particular extreme of the PF. This is because if the ROI is on one side of the PF, it is more or less close to the boundary. The baseline MOEA/D and NSGA-III, which were originally designed to approximate the whole PF, can always find solutions on the boundary, whereas it becomes increasingly difficult to find solutions on the middle region of the PF with the increase of the number of objectives. Therefore, the approximation error to a DM's golden point on one side of the PF seems to be better than those on the middle region of the PF. In contrast, since our proposed interactive framework can progressively learn the DM's preference information and adjust the search direction, I-MOEA/D-PLVF and I-NSGA-III-PLVF can well approximate the ROI in any part of the PF.

\item Furthermore, we find that the performance of I-MOEA/D-PLVF and I-NSGA-III-PLVF do not depend on the shape of the PF (in particular DTLZ1 test problem has a linear PF while DTLZ2 to DTLZ4 test problems have a concave PF). But the performance of the proposed interactive framework can be influenced by the difficulty of the search space. In particular, if the search space contains many local PFs, like DTLZ1 and DTLZ3, the evolving population may need a long time to jump over these local PFs. Even worse, some region of the PF will be more difficult to approximate than the others. If this region happens to be the ROI, the DM will wrongly assign a higher score to the solutions outside the ROI. In~\pref{fig:traj_c} and~\pref{fig:traj_b}, we plot the variations of the approximation error versus the number of generations on the 3-objective case while more comprehensive results can be found in Section IV of the supplementary document. From these plots, we can see that the approximation error on the relatively simple DTLZ2 test problem quickly drops down at the early stage of the evolution. But for problems with many local PFs, i.e., DTLZ1 and DTLZ3 test problems, the trajectories of approximation error struggle longer time before dropping down. Although DTLZ4 test problem does not have local PFs, its search space has strong bias toward certain objective coordinates. Accordingly, we observe the the fluctuation of the trajectories over generations. This might be caused by the biased evolving population which mislead the DM in decision-making.

\begin{figure*}[htbp]
\centering
\includegraphics[width=\linewidth]{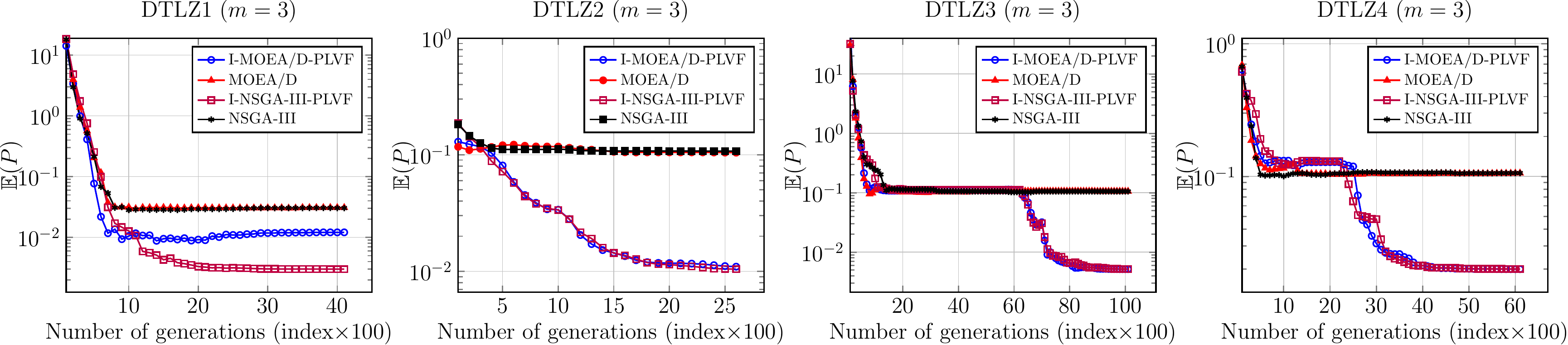}
\caption{Trajectories of the approximation error versus the number of generations on 3-objective DTLZ1 to DTLZ4 test problems. The DM’s “golden” point prefers the middle region of the PF.}

\label{fig:traj_c}
\end{figure*}

\begin{figure*}[htbp]
\centering
\includegraphics[width=\linewidth]{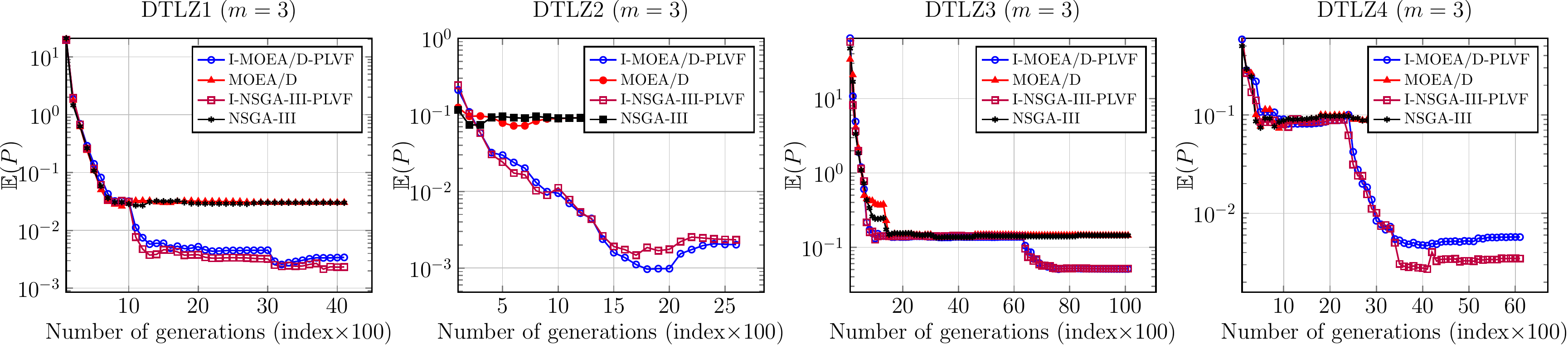}
\caption{Trajectories of the approximation error versus the number of generations on 3-objective DTLZ1 to DTLZ4 test problems. The DM’s “golden” point prefers one side of the PF.}
\label{fig:traj_b}
\end{figure*}

\end{itemize}

\subsection{Parametric Studies}
\label{sec:parametric}

As introduced in~\pref{sec:settings}, besides the intrinsic parameters associated with an EMO algorithm, e.g., population size, crossover and mutation probabilities, in our proposed interactive framework, there are some additional parameters that may affect the performance for approximating the ROI. They are: the number of incumbent candidates presented to the DM for scoring $(\mu)$, the number of generations between two consecutive consultation sessions  $(\tau)$, and the step size of the reference point update $(\eta)$. In this subsection, we study the effects of these parameters, while keeping the other parameters of I-MOEA/D-PLVF and I-NSGA-III-PLVF the same as introduced in~\pref{sec:settings}. In particular, we use DTLZ1 and DTLZ2 as the test problems, given the observations on DTLZ3 and DTLZ4 test problems can be generalized from those in DTLZ2. Each algorithm is run 21 times with different random seeds.

\subsubsection{Effect of $\mu$}
\begin{figure}[htbp]
\centering
\includegraphics[width=.7\linewidth]{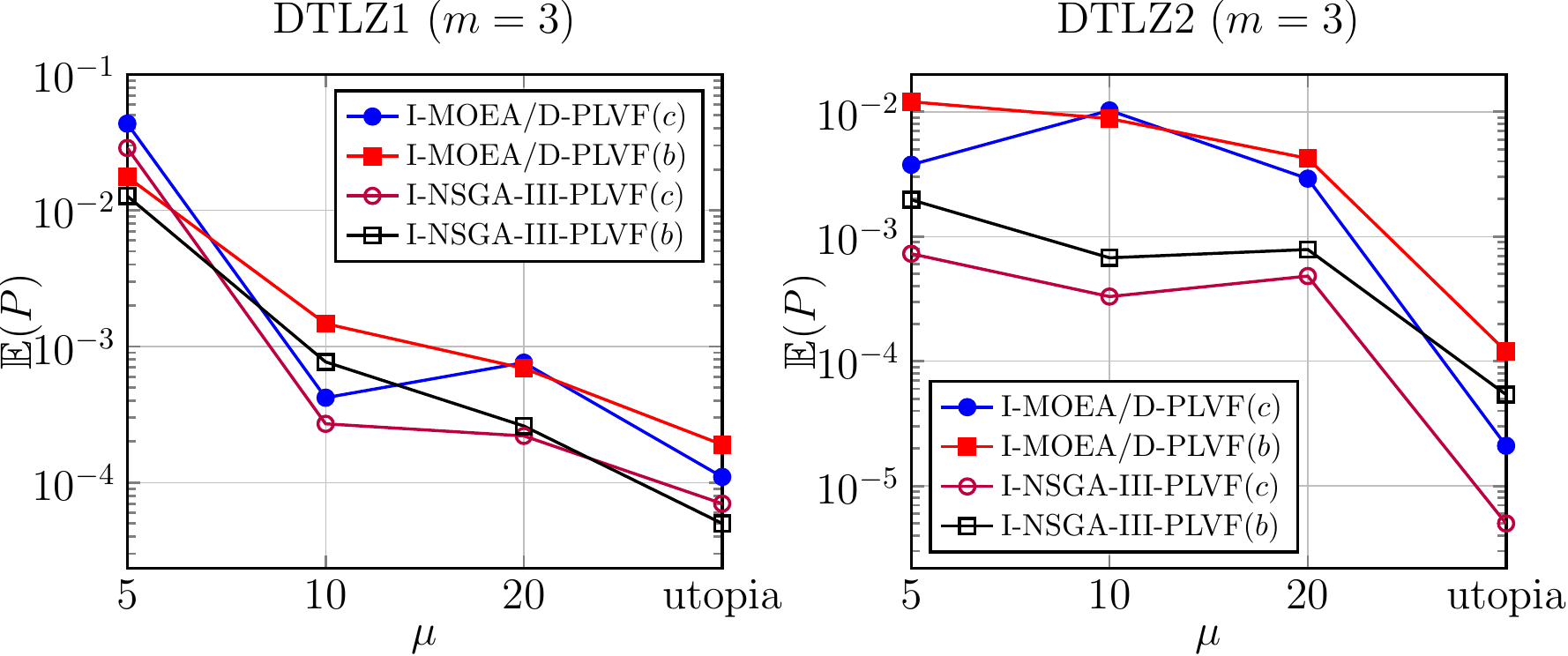}
\caption{Variations of the approximation errors with different $\mu$ settings. $(c)$ indicates the preference on the middle region of the PF, while $(b)$ indicates the preference on an extreme.}
\label{fig:mu}
\end{figure}

As introduced in~\pref{sec:consultation}, $\mu$ determines the number of labeled data (scored by the DM) that can be used to train the AVF model. It makes sense that the more data you can provide, the more accurate AVF model you can expect. However, presenting the DM too many alternatives for scoring will definitely increase her/his workload, thus lead to the fatigue. On the other hand, the model accuracy will be impaired if the data is not sufficient. To study the effect of $\mu$, we consider three different settings, i.e., $\mu\in\{5,10,20\}$. Furthermore, to validate the importance of an accurate AVF model for helping the interactive framework, we also investigate an \lq\lq utopia\rq\rq\ scenario where I-MOEA/D-PLVF and I-NSGA-III-PLVF directly use the DM's golden value function in the preference elicitation module. In~\pref{fig:mu}, we show the variations of the median approximation error with respect to different $\mu$ settings and the utopia scenario. Note that we only show the three-objective case since the other observations are similar (more comprehensive results can be found in Section V of the supplementary document). As expected, I-MOEA/D-PLVF and I-NSGA-III-PLVF always perform best when directly using the DM's golden value function. This observation supports the importance of an accurate model. Moreover, I-MOEA/D-PLVF and I-NSGA-III-PLVF can have a better performance when using a large $\mu$.

\subsubsection{Effect of $\tau$}
\begin{figure}[htbp]
\centering
\includegraphics[width=.7\linewidth]{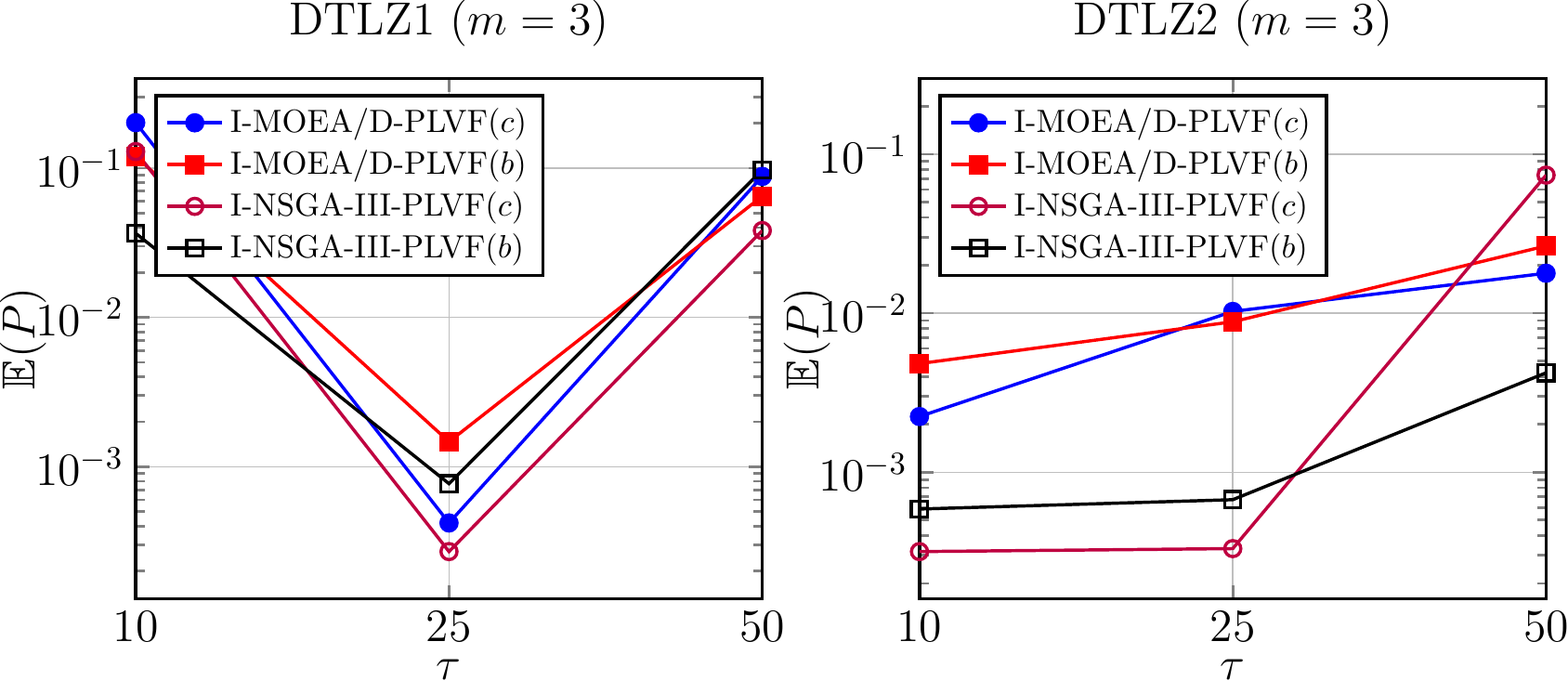}
\caption{Variations of the approximation errors with different $\tau$ settings. $(c)$ indicates the preference on the middle region of the PF, while $(b)$ indicates the preference on an extreme.}
\label{fig:tau}
\end{figure}

Here we study the effect of $\tau$ by considering three different settings, i.e, $\tau\in\{10,25,50\}$. In~\pref{fig:tau}, we plot the variations of the median approximation error with respect to different $\tau$ settings in the three-objective scenario while more comprehensive results can be found in Section V of the supplementary document. Specifically, a small $\tau$ means that we need to frequently ask the DM for scoring the candidate solutions and then update the AVF model accordingly. To a certain extent, this operation can improve the model accuracy for approximating the DM's preference information. However, similar to the overfitting phenomenon in machine learning, too frequent DM calls also have the risk of premature convergence on some local optima. As shown in~\pref{fig:tau}, the performance of I-MOEA/D-PLVF and I-NSGA-III-PLVF is not promising when setting $\tau=10$ on DTLZ1 test problem which has more than $11^4$ local PFs. On the other hand, if the DM is rarely been consulted by using a large $\tau$, the consultation module can hardly get enough information from the DM. Thus, we can hardly expect that the AVF model can provide useful information that truly represent the DM's preference information to the optimization module. As expected, the performance of I-MOEA/D-PLVF and I-NSGA-III-PLVF is always not satisfactory when setting $\tau=50$.

\subsubsection{Effect of $\eta$}
As introduced in~\pref{sec:preference}, $\eta$ controls the convergence rate of the reference points toward the promising ones identified by the AVF model learned from the consultation module. A large $\eta$ will lead to a fast convergence, thus it may have a risk of pre-mature convergence toward an undesired region. On the contrary, a small $\eta$ may slow down the convergence toward the ROI within the limited number of FEs. To study the effect of $\mu$, we consider different $\mu$ settings as $\eta\in\{0.1,0.3,0.5,0.7,0.9\}$. From the results shown in~\pref{fig:eta} and more comprehensive results shown in Section V of the supplementary document, we find that the best setting of $\eta$ is problem dependent. But neither too large nor too small $\eta$ can offer a satisfactory result.

\begin{figure}[htbp]
\centering
\includegraphics[width=.7\linewidth]{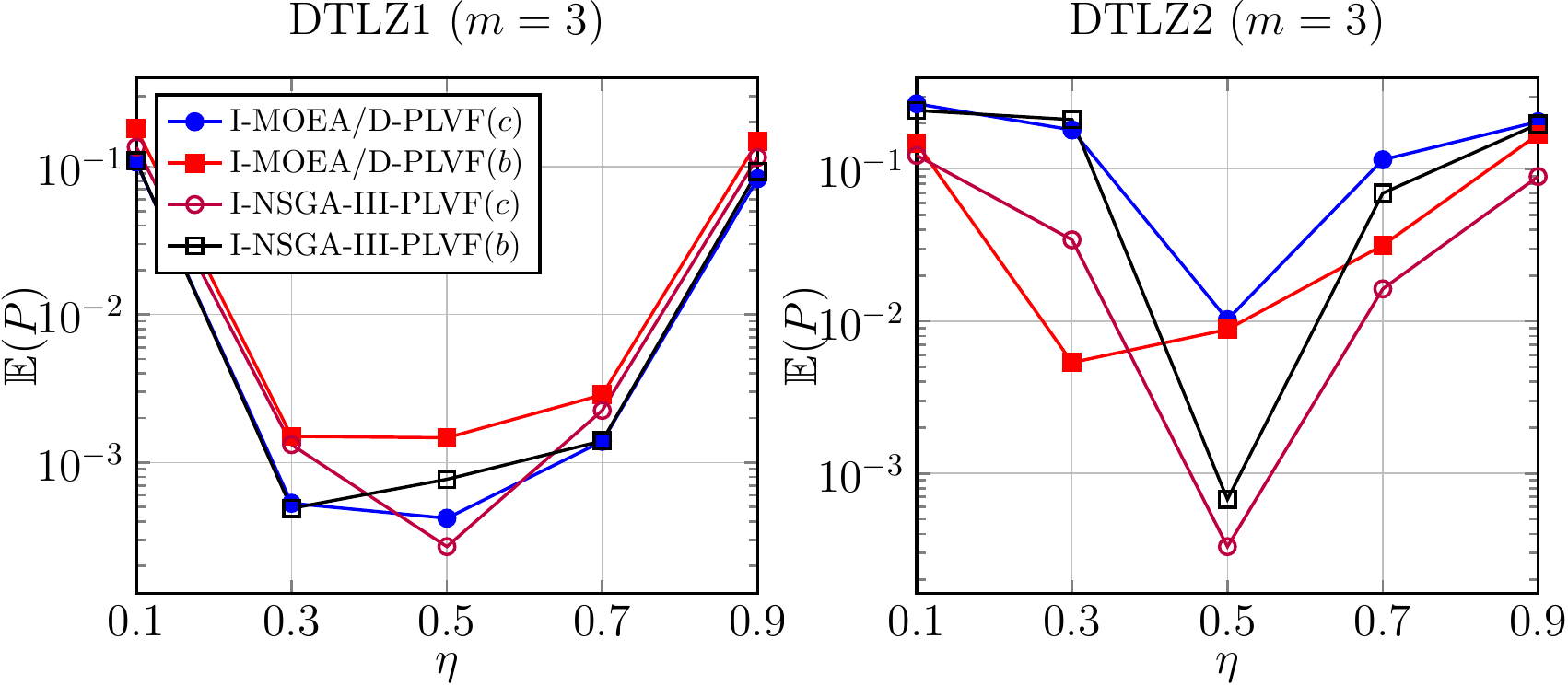}
\caption{Variations of the approximation errors with different $\eta$ settings. $(c)$ indicates the preference on the middle region of the PF, while $(b)$ indicates the preference on an extreme.}
\label{fig:eta}
\end{figure}

\subsection{Random Error in DM's Preference Information}
\label{sec:uncertainty}

\begin{figure}[htbp]
\centering
\includegraphics[width=.7\linewidth]{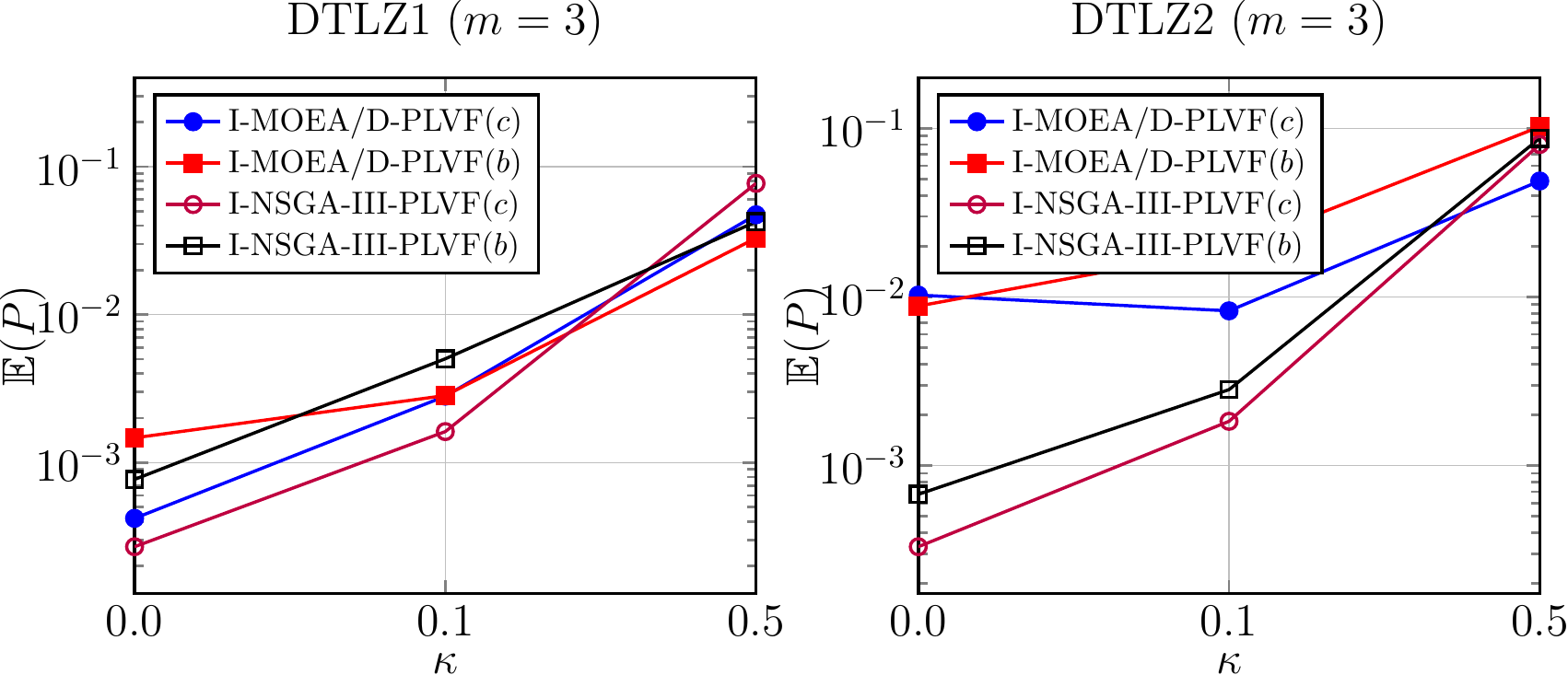}
\caption{Variations of the approximation errors with different $\kappa$ settings. $(c)$ indicates the preference on the middle region of the PF, while $(b)$ indicates the preference on an extreme.}
\label{fig:noise}
\end{figure}

In the previous experiments, the DM's preference information is modeled as a deterministic value function. However, as a human being, the DM can show certain level of inconsistencies when providing her/his preference information in practice. To study the effect brought by the inconsistencies in the preference elicitation, we consider adding some stochastic noises to the DM's golden value function. Specifically, the stochastic DM's golden value function is defined as:
\begin{equation}
    \overline{\psi}(\mathbf{x})=\psi(\mathbf{x})\times\mathsf{noise}(1,\delta_t^2)
\end{equation}
where $\psi(\mathbf{x})$ is calculated according to~\pref{eq:vf}. $\mathsf{noise}(\ast,\ast)$ is a Gaussian noise, $\delta_t=\kappa\times(1-\frac{T}{T_{\max}})$, $\kappa$ determines the strength of the noise level. $T$ is the current generation counter and $T_{\max}$ is the pre-defined maximum number of generations. In this case, we can expect that the DM's inconsistencies decrease with the evolution. This setting tries to simulate a realistic DM who is likely to make errors at the beginning due to her/his lack of knowledge of the problem. With the progress of evolution, the DM becomes more consistent in preference elicitation and decision-making. To study the influence of DM's inconsistencies on the performance of our proposed interactive framework, we set $\kappa\in\{0.0, 0.1, 0.5\}$ separately. In particular, it becomes the deterministic value function if $\kappa=0.0$. Here we choose DTLZ1 and DTLZ2 as the test problems, and keep the parameters of I-MOEA/D-PLVF and I-NSGA-III-PLVF the same as introduced in~\pref{sec:settings}. \pref{fig:noise} shows the variations of the approximation error with respect to different $\kappa$ settings (more comprehensive results are shown in Section VI of the supplementary document). As expected the performance of I-MOEA/D-PLVF and I-NSGA-III-PLVF degenerates with the increase of $\kappa$.

%% file: conclusion.tex

\section{Conclusions and Future Directions}
\label{sec:conclusions}

This paper has proposed a simple yet elegant interactive framework which can be embedded in any decomposition-based EMO algorithm so that the ultimate goal is not a complete PF, but a preferred solution of the DM's choice. This interactive framework consists of three modules, i.e, consultation, preference elicitation and optimization. Specifically, the consultation module aims to progressively learn an AVF to model the DM's preference information. In particular, during the consultation session, the DM is presented with a few incumbent candidates for scoring according her/his preference. Once the AVF is learned, the preference elicitation module translates it into the form that can be used in a decomposition-based EMO algorithm, i.e., a set of reference points that are biased toward the ROI. Extensive experiments on three to ten-objective test problems fully demonstrate the effectiveness of our proposed interactive framework for helping two widely used decomposition-based EMO algorithms, i.e., MOEA/D and NSGA-III, for finding the DM's preferred solution.

The proposed interactive framework can be extended in a number of different ways.
\begin{itemize}
\item First, this paper assume that the DM's preference information is represented as a monotonic value function. However, in practice, it is not uncommon that the DM judges some of the alternatives to be incomparable. How to discriminate the order information from incomparable comparisons? Moreover, instead of assigning a scalar score to a solution, it is interesting to study how to derive the preference information through holistic comparisons among incumbent solutions.
\item This paper only considers the unconstrained MOPs, in which any preference information elicited by the DM is feasible. In the presence of constraints or the PF by itself is discontinuous, some region of the attainable objective space is infeasible but is also unknown to the DM. It is interesting to study how to handle the infeasible information and to inform the DM to modify her/his preference information in due course.
\item Although this paper has restricted the value function to be of certain form~(\ref{eq:vf}), other more value function structures can also be considered. Furthermore, it is interesting to further investigate the robustness consideration in deriving the AVF. More studies are required to investigate the side effects brought by the inconsistencies in decision-making and the ways to mitigate that.
\item In this paper, the termination criterion is the same as the standard EMO algorithm, i.e., the fixed number of FEs. However, as shown in the plots of the trajectories of the approximation errors, we can see that algorithms can be terminated earlier before the pre-defined number of FEs. It is interesting to investigate some optimal stopping criterion that is able to automatically terminates an algorithm when it is necessary.
\end{itemize}

As discussed in~\cite{PurshouseDMMW14}, it seems to be an inspiring direction for future research of MO to combine the ideas from the EMO algorithms and MCDM. More efforts are required to develop pragmatic algorithms for MO and decision-making.